\documentclass{article}
\usepackage[dvipsnames,table,xcdraw]{xcolor}

\usepackage[sectionbib,round]{natbib}
\usepackage{arxiv}
\usepackage[utf8]{inputenc} 
\usepackage[T1]{fontenc}    
\usepackage{url}            
\usepackage{booktabs}       
\usepackage{amsfonts}       
\usepackage{nicefrac}       
\usepackage{microtype}      
\usepackage{lipsum}

\usepackage{hyperref}
\usepackage{graphicx}
\usepackage{url}
\usepackage{amssymb}
\usepackage{amsmath}
\usepackage{amsthm}
\usepackage{tabularx}
\usepackage{multicol}
\usepackage{multirow}
\usepackage{wrapfig}
\usepackage{soul}

\theoremstyle{definition}
\newtheorem{definition}{Definition}

\usepackage[margin=6pt,font={footnotesize},labelfont=bf]{caption}
\definecolor{linkcolour}{rgb}{0,0.2,0.6}
\definecolor{linkcolour2}{rgb}{0.7,0.1,0.1}
\hypersetup{colorlinks,breaklinks,linkcolor=linkcolour2,citecolor=linkcolour,filecolor=linkcolour,urlcolor=linkcolour}

\usepackage{titlesec}
\titlespacing*{\section}{0pt}{0.5\baselineskip}{0.1\baselineskip}
\titlespacing*{\subsection}{0pt}{0.3\baselineskip}{0.1\baselineskip}
\titlespacing*{\subsubsection}{0pt}{0.1\baselineskip}{0.1\baselineskip}

\title{\bf Hyper-SAGNN: a self-attention based graph neural network\\ for hypergraphs}

\author{
Ruochi Zhang\\
School of Computer Science\\
Carnegie Mellon University 
\And
Yuesong Zou\\
School of Computer Science\\
Carnegie Mellon University\\
IIIS, Tsinghua University 
\And
Jian Ma\\
School of Computer Science\\
Carnegie Mellon University\\
\texttt{jianma@cs.cmu.edu}
}

\begin{document}

\maketitle

\begin{abstract}
Graph representation learning for hypergraphs can be used to extract patterns among higher-order interactions that are critically important in many real world problems.
Current approaches designed for hypergraphs, however, are unable to handle different types of hypergraphs and are typically not generic for various learning tasks.
Indeed, models that can predict variable-sized heterogeneous hyperedges have not been available.
Here we develop a new self-attention based graph neural network called Hyper-SAGNN applicable to homogeneous and heterogeneous hypergraphs with variable hyperedge sizes. 
We perform extensive evaluations on multiple datasets, including four benchmark network datasets and two single-cell Hi-C datasets in genomics.
We demonstrate that Hyper-SAGNN significantly outperforms the state-of-the-art methods on traditional tasks while also achieving great performance on a new task called outsider identification.
Hyper-SAGNN will be useful for graph representation learning to uncover complex higher-order interactions in different applications. 
\end{abstract}

\vspace{15pt}

\section{Introduction}

\begin{wrapfigure}{r}{0.35\textwidth}
    \centering
    \includegraphics[width=0.35\textwidth]{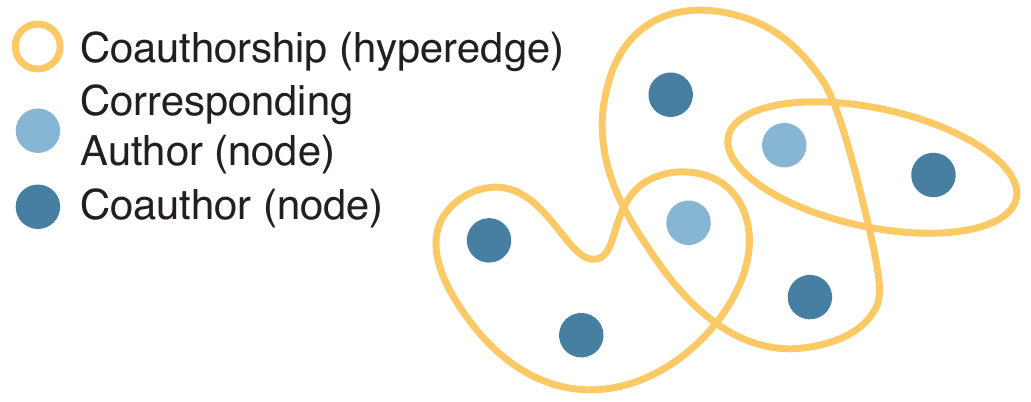}
    \caption{An example of the co-authorship hypergraph. 
    Here authors are represented as nodes (in dark blue and light blue) and coauthorships are represented as hyperedges.}
    \label{fig:intro}
\end{wrapfigure}

Graph structure is a widely used representation for data with complex interactions. 
Learning on graphs has also been an active research area in machine learning on how to predict or discover patterns based on the graph structure~\citep{hamilton2017representation}.
Although existing methods can achieve strong performance in tasks such as link prediction and node classification, they are mostly designed for analyzing pair-wise interactions and thus are unable to effectively capture higher-order interactions in graphs. 
In many real-world applications, however, relationships among multiple instances are key to capturing critical properties, 
e.g., co-authorship involving more than two authors
or relationships among multiple heterogeneous objects such as ``(human, location, activity)''.
Hypergraphs can be used to represent higher-order interactions~\citep{zhou2007learning}.
To analyze higher-order interaction data, it is straightforward to expand each hyperedge into pair-wise edges with the assumption that the hyperedge is decomposable. 
Several previous methods were developed based on this notion~\citep{sun2008hypergraph,feng2018learning}. 
However, earlier work DHNE (Deep Hyper-Network Embedding)~\citep{tu2018structural} suggested the existence of heterogeneous indecomposable hyperedges where relationships within an incomplete subset of a hyperedge do not exist. 
Although DHNE provides a potential solution by modeling the hyperedge directly without decomposing it, 
due to the neural network structure used in DHNE, the method is limited to the fixed type and fixed-size heterogeneous hyperedges and is unable to consider relationships among multiple types of instances with variable size.
For example, Fig.~\ref{fig:intro} shows a heterogeneous co-authorship hypergraph with two types of nodes (corresponding author and coauthor). 
Due to the variable number of both authors and corresponding authors in a publication, the hyperedges (co-authorship) have different sizes or types.
Unfortunately, methods for representation learning of heterogeneous hypergraph with variable-sized hyperedges, especially those that can predict variable-sized hyperedges, have not been developed.

In this work, we developed a new self-attention based graph neural network, called Hyper-SAGNN that can work with both homogeneous and heterogeneous hypergraphs with variable hyperedge size. 
Using the same datasets in the DHNE paper~\citep{tu2018structural}, we demonstrated the advantage of Hyper-SAGNN over DHNE in multiple tasks.
We further tested the effectiveness of the method in predicting edges and hyperedges and showed that the model can achieve better performance from the multi-tasking setting.
We also formulated a novel task called outsider identification and showed that Hyper-SAGNN performs strongly.
Importantly, as an application of Hyper-SAGNN to single-cell genomics, we were able to learn the embeddings for the most recently produced single-cell Hi-C (scHi-C) datasets to uncover the clustering of cells based on their 3D genome structure~\citep{ramani2017massively,nagano2017cell}.
We showed that Hyper-SAGNN achieved improved results in identifying distinct cell populations as compared to existing scHi-C clustering methods.
Taken together, Hyper-SAGNN can significantly outperform the state-of-the-art methods and can be applied to a wide range of hypergraphs for different applications.

\section{Related Work}
Deep learning based models have been developed recently to generalize from graphs to hypergraphs~\citep{gui2016large,tu2018structural}. 
The HyperEdge Based Embedding (HEBE) method~\citep{gui2016large} aims to learn the embeddings for each object in a specific heterogeneous event by representing it as a hyperedge. 
However, as demonstrated in~\citet{tu2018structural}, HEBE does not perform well on sparse hypergraphs.
Notably, previous methods typically decompose the hyperedge into pair-wise relationships where the decomposition methods can be divided into two categories: 
explicit and implicit.
For instance, given a hyperedge $(v_1,v_2,v_3)$, the explicit approach would decompose it directly into three edges,
$(v_1,v_2),(v_2,v_3),(v_1,v_3)$, while the implicit approach would add a hidden node $e$ representing the hyperedge before decomposition, i.e., $(v_1,e),(v_2,e),(v_3,e)$.
The deep hypergraph embedding (DHNE) model, however, directly models the tuple-wise relationship using MLP (Multilayer Perceptron).
The method is able to achieve better performance on multiple tasks as compared to other methods designed for graphs or hypergraphs such as Deepwalk~\citep{perozzi2014deepwalk}, node2vec~\citep{grover2016node2vec}, and HEBE.
Unfortunately, the structure of MLP takes fixed-size input, making DHNE only capable of handling $k$-uniform hypergraphs, i.e., hyperedges containing $k$ nodes.
To use DHNE for non-$k$-uniform hypergraphs or hypergraphs with different types of hyperedges, a function for each type of hyperedges needs to be trained individually, 
which leads to significant computational cost and loss of the capability to generalize to unseen types of hyperedges.
Another recent method, hyper2vec~\citep{huang2019hyper2vec}, can also generate embeddings for nodes within the hypergraph.
However, hyper2vec cannot solve the link prediction problem directly as it only generates the embeddings of nodes in an unsupervised manner without a learned function to map from embeddings of nodes to hyperedges.
Also, for $k$-uniform hypergraphs, hyper2vec is equivalent to node2vec, which cannot capture the high-order network structures for indecomposable hyperedges (as shown in~\citet{tu2018structural}). 
Our Hyper-SAGNN in this work addresses all these challenges with a self-attention based graph neural network that can learn embeddings of the nodes and predict hyperedges for non-$k$-uniform heterogeneous hypergraphs.

\section{Method}
\subsection{Definitions and Notations}
\begin{definition}
\textbf{(Hypergraph)} A hypergraph is defined as $G = (V,E)$, where $V = \{v_1,...,v_n\}$ represents the set of nodes in the graph, and $E = \{e_i = (v_1^{(i)},...,v_k^{(i)})\}$ represents the set of hyperedges. 
For any hyperedge $e$, it can contain more than two nodes (i.e., $\delta(e) \geq 2$).
If all hyperedges within a hypergraph have the same size of $k$, it is called a $k$-uniform hypergraph.
Note that even if a hypergraph is $k$-uniform, it can still have different types of hyperedges because the node type can vary for nodes within the hyperedges. 
\end{definition}

\begin{definition}
\textbf{(The hyperedge prediction problem)}
We formally define the hyperedge prediction problem. 
For a given tuple $(v_1, v_2, ..., v_k)$, our goal is to learn a function $f$ that satisfies:
\begin{align}
    f(v_1, v_2, ..., v_k) &= \left\{ \begin{array}{ll}
    \geq s, &\text{if}~(v_1, v_2, ..., v_k) \in E \\
     < s, &\text{if}~(v_1, v_2, ..., v_k) \notin E
     \end{array} \right. 
\end{align}
where $s$ is the threshold to binarize the continuous value of $f$ into a label, which indicates whether the tuple is an hyperedge or not. 
Specifically, when we are given the pre-trained embedding vectors or the features of nodes $X = \{x_1,...,x_i\}$, we can rewrite this function as:
\begin{align}
    f(v_1, v_2, ..., v_k) &\triangleq f(g(x_1),g(x_2),...,g(x_k))
\end{align}
where the vectors $g(x_i)$ can be considered as the fine-tuned embedding or embedding vectors for the nodes. 
For convenience, we refer to $x_i$ as the features and $g(x_i)$ as the learned embeddings.
\end{definition}

\subsection{Structure of Hyper-SAGNN}

Our goal is to learn the functions $f$ and $g$ that take tuples of node features $(x_1,..., x_k)$ as input and produce the probability of these nodes forming a hyperedge. 
Without the assumption that the hypergraph is $k$-uniform and the type of each hyperedge is identical, we require that $f$ can take variable-sized, non-ordered input. 
Although simple functions such as average pooling $f(g(x_1), ..., g(x_k)) = \frac{1}{K} \sum_{i=1}^k g(x_i)$ satisfy this tuple-wise condition, previous work showed that the linear function is not sufficient to model this relationship~\citep{tu2018structural}. 
DHNE used an MLP to model the non-linear function, but it requires that an individual function needs to be trained for different types of hyperedges. 
Here we propose a new method to tackle the general hyperedge prediction problem.

\begin{figure}[ht!]
    \centering
    \includegraphics[width=0.7\textwidth]{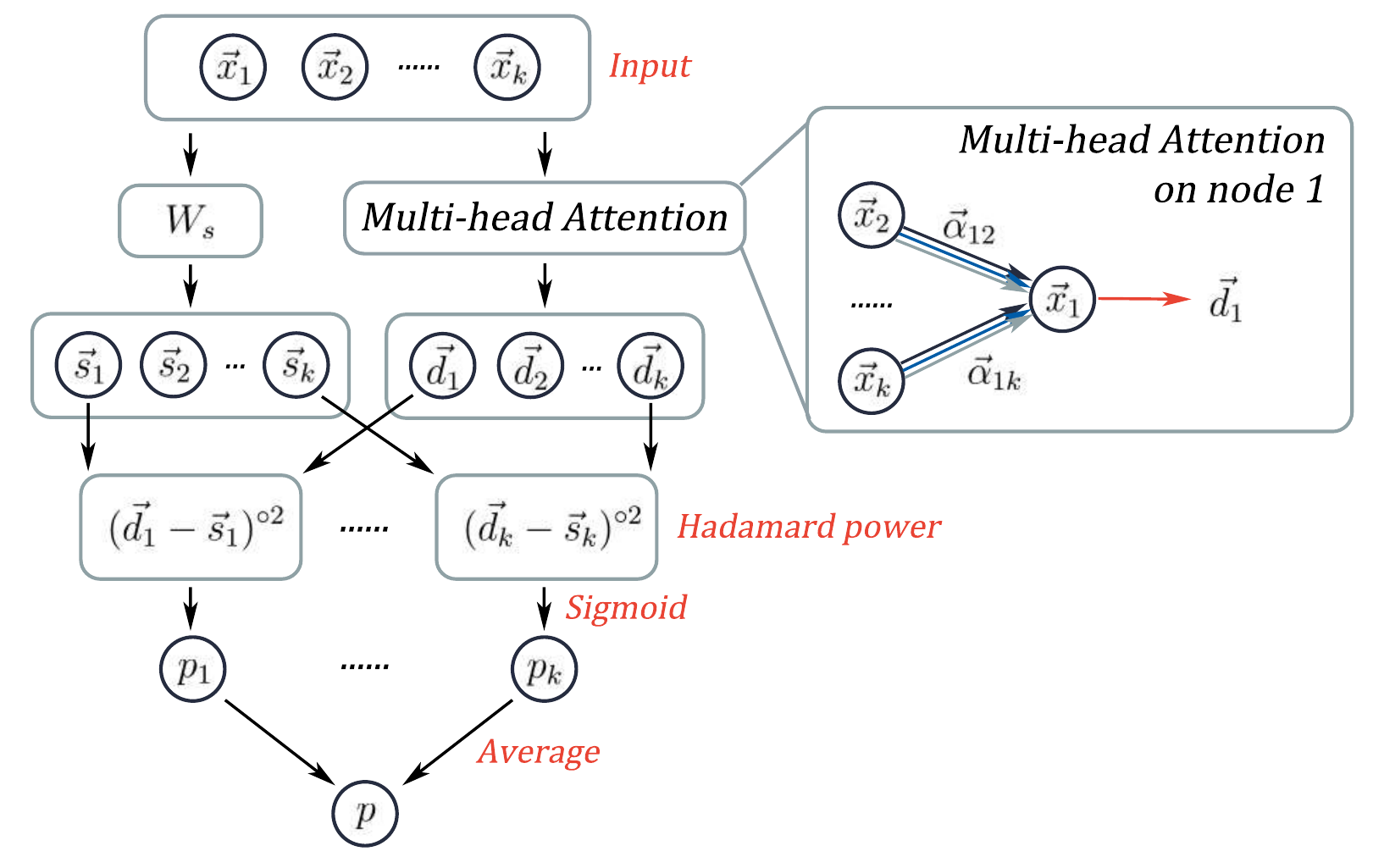}
    \caption{Structure of the neural network used in Hyper-SAGNN.
    The input $(\vec{x}_1, \vec{x}_2,...,\vec{x}_k)$, representing the features for nodes 1 to $k$, passes through two branches of the network resulting in static embeddings $(\vec{s}_1, \vec{s}_2,...,\vec{s}_k)$ and dynamic embeddings $(\vec{d}_1, \vec{d}_2,...,\vec{d}_k)$, respectively. 
    The layer for generating dynamic embeddings is the multi-head attention layer. 
    An example for its mechanism on node 1 here is shown in the figure as well.
    Then the pseudo-euclidean distance of each pair of static and dynamic embeddings is calculated by one-layered position-wise feed-forward network to produce probability scores $(p_1,p_2,...,p_k)$. 
   These scores are further averaged to represent whether this group of nodes form a hyperedge or not.}
    \label{fig:struct}
\end{figure}

Graph neural network based methods such as GraphSAGE~\citep{hamilton2017inductive} typically define a unique computational graph 
for each node, allowing it to perform efficient information aggregation for nodes with different degrees.
Graph Attention Network (GAT) introduced by~\citet{velivckovic2017graph} utilizes a self-attention mechanism in the information aggregation process.
Motivated by these properties, we propose our method Hyper-SAGNN based on self-attention mechanism within each tuple to learn the function $f$. 

We first briefly introduce the self-attention mechanism.
We use the same terms as the self-attention mechanism described in~\citet{vaswani2017attention, velivckovic2017graph}.
Given a group of nodes $(\vec{x}_1,\vec{x}_2,...,\vec{x}_k)$ and weight matrices $W_Q, W_K, W_V$ that represent linear transformation of features before applying the dot-product attention to be trained, we first compute the attention coefficients that reflect the pair-wise importance of nodes:
\begin{align}
    e_{ij} = \left (W_Q^Tx_i\right )^T\left (W_K^Tx_j\right ), \forall 1\leq i,j \leq k
\end{align}
We then normalize $e_{ij}$ by all possible $j$ within the tuple through the softmax function, i.e., 
\begin{align}
    \alpha_{ij} = \frac{\exp(e_{ij})}{\sum_{1 \leq l \leq k} \exp(e_{il})}
\end{align}
Finally, a weighted sum of the transformed features with an activation function is calculated:
\begin{align}
    \vec{d}_i = \tanh\left (\sum_{1\leq j \leq k} \alpha_{ij} W_V^T x_j \right ) \label{eq:eq1}
\end{align}

In GAT, each node is applied to the self-attention mechanism usually with all its first-order neighbors. 
In Hyper-SAGNN, we aggregate the information for a node $v_i$ only with its neighbors for a given tuple. 
The structure of Hyper-SAGNN is illustrated in Fig.~\ref{fig:struct}. 

The input to our model can be represented as tuples, i.e., $(\vec{x}_1,\vec{x}_2,...,\vec{x}_k)$. 
Each tuple first passes through a position-wise feed-forward network to produce $(\vec{s}_1,\vec{s}_2,...,\vec{s}_k)$, where $\vec{s}_i = \tanh(W_s^T \vec{x}_i)$. 
We refer to each $\vec{s}_i$ as the static embedding for node $i$ since it remains the same for node $i$ no matter what the given tuple is.
The tuple also passes through a multi-head graph attention layer to produce a new set of node embedding vectors $(\vec{d}_1,\vec{d}_2,...,\vec{d}_k)$, which we refer to as the dynamic embeddings because they are dependent on all the node features within this tuple.

Note that unlike the standard attention mechanism described above, when calculating $\vec{d}_i$, we require that $j \neq i$ in Eqn.~(\ref{eq:eq1}).
In other words, we exclude the term $\alpha_{ii}W_V^Tx_i$ in the calculation of dynamic embeddings.
Based on our results we found that including $\vec{\alpha}_{ii}$ would lead to either similar or worse performance in terms of the hyperedge prediction and node classification task
(see Appendix~\ref{app:variants} for details).
We will elaborate on the motivation of this choice later in this section.

With the static and dynamic embedding vectors for each node, we calculate the Hadamard power (element-wise power) of the difference of the corresponding static/dynamic pair.
It is then further passed through a one-layered neural network with sigmoid as the activation function to produce a probability score $p_i$.
Finally, all the output $p_i \in [0,1]$ is averaged to get the final $p$, i.e.,
\begin{align}
    &o_i = W_o^T ((\vec{d}_i - \vec{s}_i)^{\circ 2}) + b \\
    &p = \frac{1}{K} \sum_{i=1}^k p_i = \frac{1}{K} \sum_{i=1}^k \sigma(o_i)
    \label{eq:p_i}
\end{align}

By design, $o_i$ can be regarded as the squared weighted pseudo-euclidean distance between the static embedding $\vec{s}_i$ and the dynamic one $\vec{d}_i$.
It is called pseudo-euclidean distance because we do not require the weight to be non-zero or to sum up to 1.
One rationale for allowing negative weights when calculating the distance could be the Minkowski space where the distance is defined as $d^2 = x^2 + y^2 + z^2 - t^2$.
Therefore, for these high dimensional embedding vectors, we do not specifically treat them as euclidean vectors.

\begin{figure}[ht!]
    \centering
    \includegraphics[width = 0.6\textwidth]{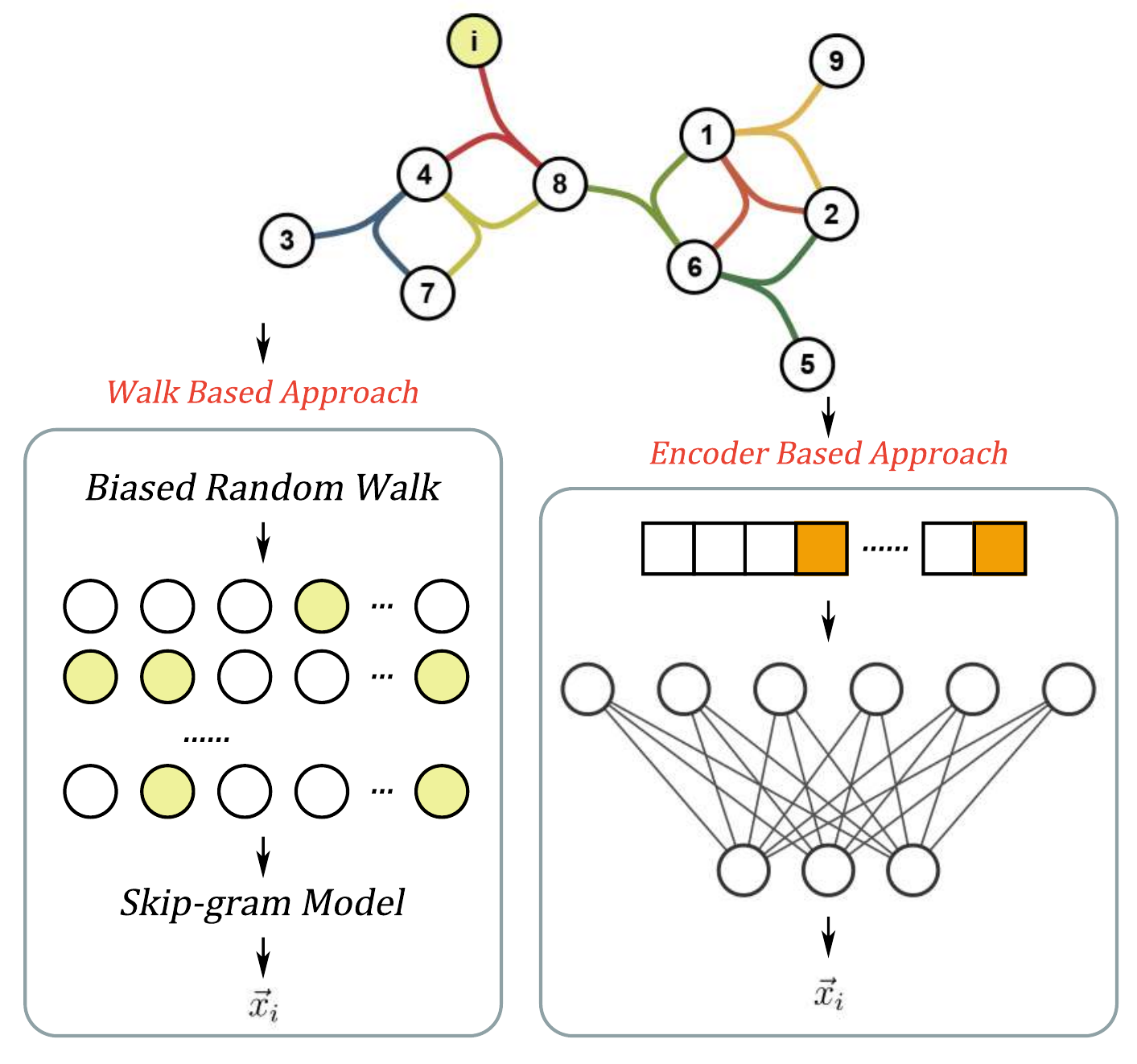}
    \caption{Illustration of the method for generating node features for node $i$ in the hypergraph. 
    In the walk based approach, a biased random walk on hypergraphs is used to produce walking paths (the yellow circles in the walking paths represents node $i$). 
    These walks are further used to train a skip-gram model for features. 
    In the encoder based approach, the $i$-th row of the adjacency matrix (as shown in the figure where the orange/white blocks represent whether or not there is adjacency between node $i$ and other nodes in the graph) is used as the input to an auto-encoder.
    The output of the encoder part is used as the features for node $i$. }
    \label{fig:x_i}
\end{figure}

Our network essentially aims to build the correlation of the average ``distance'' of the static/dynamic embedding pairs with the probability of the node group forming a hyperedge.
Since the dynamic embedding is the weighted sum of features (with potential non-linear transformation) from neighbors within the tuple, this ``distance'' reflects how well the static embedding of each node can be approximated by the features of their neighbors within that tuple.
This design strategy shares some similarities with the CBOW model in natural language processing~\citep{mikolov2013skipgram}, where the model aims to predict the target word given its context.
In principle, we could still include the $\vec{\alpha}_{ii}$ term to obtain the embedding $\vec{d}^{*}_i$.
Alternatively, we can directly pass $\vec{d}^{*}_i$ through a fully connected layer to produce $p^{*}_i$ while the rest remains the same.
However, we argue that our proposed model would be able to produce $s_i$ that can be directly used for tasks such as node classification while the alternative approach is unable to achieve that
(see Appendix~\ref{app:variants} for detailed analysis).

\subsection{Approaches for Generating Features}

In an inductive learning setting with known attributes for the nodes, $\vec{x}_i$ can just be the attributes of the node. 
However, in a transductive learning setting without knowing attributes of the nodes, we have to generate $\vec{x}_i$ based on the graph structure solely.
Here we use two existing strategies to generate features $\vec{x}_i$.

We first define the functions used in the subsequent sections as follows: 
a hyperedge $e$ with weight $w(e)$ is incident with a vertex $v$ if and only if $v\in e$.
We denote the indicator function that represents the incident relationship between $v$ and $e$ by $h(v,e)$, which equals $1$ when $e$ is incident with $v$.
The degree of vertex, $d(v)$, and the size of hyperedge, $\delta(e)$, are defined as:
\begin{align}
    d(v) &\triangleq \sum_{e\in E} h(v,e) w(e) \\
    \delta(e) &\triangleq \sum_{v\in V} h(v,e) = |e|
\end{align}

\subsubsection{Encoder based approach}

As shown on the right side of Fig.~\ref{fig:x_i}, the first method to generate features is referred to as the encoder based approach, which is similar to the structure used in DHNE~\citep{tu2018structural}.
We first obtain the incident matrix of the hypergraph $H \in \mathbb{R}^{|V|\times |E|}$ with entries $h(v,e) = 1$ if $v\in e$ and 0 otherwise. 
We also calculate the diagonal degree matrix $D_v$ containing the vertex degree $d(v)=\sum_{e\in E} h(v,e)$.
We thus have the adjacency matrix $A=HH^T-D_v$, of which the entries $a(v_i,v_j)$ denote the concurrent times between each pair of nodes $(v_i,v_j)$.
The $i$-th row of $A$, denoted by $\vec{a}_i$, shows the neighborhood structures of the node $v_i$, which then passes through a one-layer neural network to produce $x_i$:
\begin{align}
    \vec{x}_i &= \tanh \left (W_{enc} \cdot \vec{a}_i + \vec{b}_{enc} \right ) 
\end{align}
In DHNE, a symmetric structure was introduced where there are corresponding decoders to transform the $\vec{x}_i$ back to $\vec{a}_i$.
\citet{tu2018structural} remarked that including this reconstruction error term would help DHNE to learn the graph structure better.
We also include the reconstruction error term into the loss function, but with tied-weights between encoder and decoder to reduce the number of parameters that need to be trained.
 
\subsubsection{Random walk based approach}

Besides the encoder based approach, we also utilize a random walk based framework to generate the feature vectors $\vec{x}_i$ (shown on the left side of Fig.~\ref{fig:x_i})
We extend the biased 2nd-order random walks proposed in node2vec~\citep{grover2016node2vec} to generalize to hypergraphs.
For a walk from $v$ to $x$ then to $t$, the strategies are described as follows.

The 1st-order random walk strategy given the current vertex $x$ is to randomly select a hyperedge $e$ incident with $x$ based on the weight of $e$ and then to choose the next vertex $y$ from $e$ uniformly~\citep{zhou2007learning}.
Therefore, the 1st-order transition probability is defined as: 
\begin{align}
    \pi_1(t|x) &\triangleq \sum_{e\in E} w(e) \frac{ h(t,e)h(x,e)}{\delta(e)}
\end{align}
We then generalize the 2nd-order bias $\alpha_{pq}$ from ordinary graph to hypergraph for a walk from $v$ to $x$ to $t$ as:
\begin{align}
    \alpha_{p,q}(t, v) = \left\{ \begin{array}{ll}
    1/p, & \mathrm{if} \;\exists e\in E, \;\mathrm{s.t.\;} t,v,x \in e \\
    1,           & \textrm{else if} \; \exists e\in E, \;\mathrm{s.t.\;} t,x \in e\\
    1/q, & \mathrm{otherwise}
    \end{array} \right. 
\end{align}
where the parameters $p$ and $q$ are to control the tendencies that encourage outward exploration and obtain a local view. 

Next we add the above terms to set the biased 2nd-order transition probability as:
\begin{align}
    \pi(t|v,x)=\left\{ \begin{array}{ll} 
    \frac{\pi_1(t|x) \cdot \alpha_{pq}(t,v) } {Z}, & \mathrm{if} \; \exists e\in E, \; \mathrm{s.t.} \; v, x \in e \\
    0 , & \mathrm{otherwise} 
    \end{array}\right.
\end{align}
where $Z$ is a normalizing factor. 

With the well-defined 2nd-order transition probability $\pi(t|v,x)$, we simulate a random walk of fixed length $l$ through a 2nd-order Markov process marked by $P(c_i=t| c_{i-1}=x, c_{i-2}=v) = \pi(t|v, x)$, where $c_i$ is the $i$-th node in the walk. 
A Skip-gram model~\citep{mikolov2013word2vec, mikolov2013skipgram} is then used to extract the node features from sampled walks such that the nodes that appear in similar contexts would have similar embeddings.

\section{Results}

\subsection{Evaluation Datasets}

We sought to compare Hyper-SAGNN with the state-of-the-art method DHNE as it has already been demonstrated with superior performance over previous algorithms such as DeepWalk, LINE, and HEBE.  
We also did not compare our Hyper-SAGNN with hyper2vec~\citep{huang2019hyper2vec} for the following reasons: (1) hyper2vec cannot be directly used for the hyperedge prediction task; and (2) for a $k$-uniform hypergraphs like the four datasets used in DHNE or the IMDb dataset used in the hyper2vec paper~\citep{huang2019hyper2vec}, it is equivalent to the standard node2vec. 

We first used the same four datasets in the original DHNE paper to have a direct comparison: 
\begin{itemize} 
\item GPS~\citep{datasetGPS}: GPS network. 
The hyperedges are based on (user, location, activity) relations. 
\item MovieLens~\citep{datasetMovieLens}: Social network. 
The hyperedges are based on (user, movie, tag) relations, describing peoples' tagging activities. 
\item drug: Medicine network from FAERS\footnote{http://www.fda.gov/Drugs/}. 
The hyperedges are based on (user, drug, reaction) relations.
\item wordnet~\citep{datasetwordnet}: Semantic network from WordNet 3.0. 
The hyperedges are based on (head entity, relation, tail entity), expressing the relationships between words. 
\end{itemize}
Details of the datasets, including node types, the number of nodes, and the number of edges, are shown in Table~\ref{4datasets}. 

\begin{table}[ht!]
\caption{Network datasets used for evaluation. 
The columns under ``\#(V)'' correspond to the columns under ``Node Type'' for each dataset.}
\label{4datasets}
\begin{center} \sf \footnotesize 
\begin{tabular}{l|lll|lll|l}
    \toprule
\textbf{Datasets} & \multicolumn{3}{c|}{\textbf{Node Type}} & \multicolumn{3}{c|}{\textbf{\#(V)}} & \textbf{\#(E)}
\\ \hline
GPS & user & location & activity & 146 & 70 & 5 & 1,436 \\
    MovieLens & user & movie & tag & 2,113 & 5,908 & 9,079 & 47,957 \\
    drug & user & drug & reaction & 12 & 1,076 & 6,398 & 171,756 \\
    wordnet & head & relation & tail & 40,504 & 18 & 40,551 & 145,966 \\
        \bottomrule
\end{tabular}
\end{center}
\end{table}

\subsection{Parameter Setting}

In this section, we describe details of the parameters used for both Hyper-SAGNN and other methods in the evaluation.
We downloaded the source code of DHNE from its GitHub repository. 
The structure of the neural network of DHNE was set to be the same as what the authors described in~\citet{tu2018structural}. 
We tuned parameters such as the $\alpha$ term and the learning rate following the same procedure. 
We also tried adding dropout between representation vectors and the fully connected layer for better performance of DHNE. 
All these parameters were tuned until it was able to replicate or even improve the performance reported in the original paper. 
To make a fair comparison, for all the results below, we made sure that the training and validation data setups were the same across different methods.

For node2vec, we decomposed the hypergraph into pairwise edges and ran node2vec on the decomposed graph. 
For the hyperedge prediction task, we first used the learned embedding to predict pairwise edges.
We then used the mean or min of the pairwise similarity as the probability for the tuple to form a hyperedge. 
We set the window size to 10, walk length to 40, the number of walks per vertex to 10, which are the same parameters used in DHNE for node2vec.
However, we found that for the baseline method node2vec, when we tuned the hyper-parameter $p,q$ and also used larger walk length, window size and walks per vertex (120, 20, 80 instead of 40, 10, 10), it would achieve comparable performance for node classification task as DHNE. 
This observation is consistent with our designed biased hypergraph random walk.
But this would result in a longer time for sampling the walks and training the skip-gram model.
We therefore kept the parameters consistent with what was used in DHNE paper. 

For our Hyper-SAGNN, we set the representation size to 64, which is the same as DHNE. 
When using the encoder based approach to calculate $x_i$, we set the encoder structure to be the same as the encoder part in DHNE. 
When using the random walk based approach, we decomposed the hypergraph into a graph as described above.
We set the window size to 10, walk length to 40, the number of walks per vertex to 10, to allow time-efficient generation of feature vector $\vec{x}_i$.
The results in Section~\ref{sec:eva} showed that even when the pre-trained embeddings are not so ideal, Hyper-SAGNN can still well capture the structure of the graph.

\subsection{Performance Comparison with Existing Methods}
\label{sec:eva}

We evaluated the effectiveness of our embedding vectors and the learned function with the network reconstruction task. 
We compared our Hyper-SAGNN using the encoder based approach and also the model using the random walk based pre-trained embeddings against DHNE and the baseline node2vec. 
We first trained the model and then used the learned embeddings to predict the hyperedge of the original network. 
We sampled the negative samples to be 5 times the amount of the positive samples following the same setup of DHNE. 
We evaluated the performance based on both the AUROC score and the AUPR score. 
\begin{table}[ht!]
\caption{AUC and AUPR values for network reconstruction. 
Model trained with the random walk based approach and the encoder based approach is marked as Hyper-SAGNN-W and Hyper-SAGNN-E, respectively.}
\begin{center}\sf \footnotesize 
\begin{tabular}{l|ll|ll|ll|ll}
\toprule
 & \multicolumn{2}{c|}{\textbf{GPS}} &  \multicolumn{2}{c|}{\textbf{MovieLens}} & \multicolumn{2}{c|}{\textbf{drug}} &  \multicolumn{2}{c}{\textbf{wordnet}} \\ 
  & AUC & AUPR & AUC & AUPR & AUC & AUPR & AUC & AUPR \\ \hline 
node2vec-mean & 0.572 & 0.188 & 0.557 & 0.197 &  0.668 & 0.246 & 0.613 & 0.215 \\
node2vec-min & 0.570 & 0.187& 0.535 & 0.186 &  0.682 & 0.257 & 0.576 & 0.201  \\ 
DHNE  &0.959 &0.836 & 0.974 & 0.878 & 0.952 & 0.873 & 0.989 & 0.953 \\ 
Hyper-SAGNN-E & 0.971 & \textbf{0.877} & 0.991 & 0.952 & 0.977 & 0.916 & 0.989 & 0.950\\ 
Hyper-SAGNN-W &\textbf{0.976} & 0.857 & \textbf{0.998} & \textbf{0.986} & \textbf{0.988} &\textbf{0.945} & \textbf{0.994} & \textbf{0.956}\\
\bottomrule 
\end{tabular} 
\end{center}
\label{tab:recon}
\end{table}

As shown in Table~\ref{tab:recon}, Hyper-SAGNN can capture the network structure better than DHNE over all datasets either using the encoder based approach or the random walk based approach.  

\begin{figure}[ht!]
    \centering
    \includegraphics[width=0.95\textwidth]{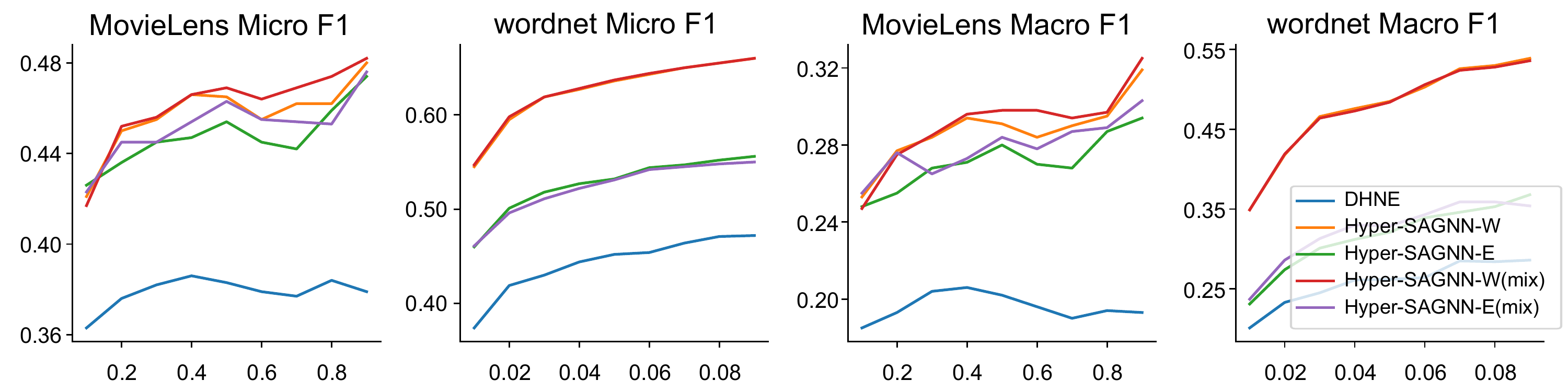}
    \caption{Performance of classification on MovieLens and wordnet datasets. 
    Hyper-SAGNN trained with the random walk based approach and encoder based approach are marked as Hyper-SAGNN-W, Hyper-SAGNN-E, respectively. 
    The models trained with a mix of edges and hyperedges are denoted with ``(mix)''.}
    \label{fig:perform_node}
\end{figure}

We further assessed the performance of Hyper-SAGNN on the hyperedge prediction task.
We randomly split the hyperedge set into training and testing set by a ratio of 4:1.
The way to generate negative samples is the same as the network reconstruction task.
As shown in Table~\ref{tab:link}, our model again achieves significant improvement over DHNE for predicting the unseen hyperedges. 
The most significant improvement is from the wordnet dataset, which is about a 24.6\% increase on the AUPR score.
For network reconstruction and hyperedge prediction tasks, the difference between random walk based Hyper-SAGNN and encoder based Hyper-SAGNN is minor.

In addition to the tasks related to the prediction of hyperedges, we also evaluated whether the learned node embeddings are effective for node classification tasks. 
A multi-label classification experiment and a multi-class classification experiment were carried out for the MovieLens dataset and the wordnet dataset, respectively. 
We used Logistic Regression as the classifier. 
The proportion of the training data was chosen to be from 10\% to 90\% for the MovieLens dataset, and 1\% to 10\% for the wordnet dataset. 
We used averaged Mirco-F1 and Macro-F1 to evaluate the performance. 
The results are in Fig.~\ref{fig:perform_node}. 
We observed that Hyper-SAGNN consistently achieves both higher Micro-F1 and Macro-F1 scores over DHNE for different fractions of the training data.
Also, Hyper-SAGNN based on the random walk generally achieves the best performance (Hyper-SAGNN-W in Fig.~\ref{fig:perform_node}).

\subsection{Performance on Non-\textit{k}-uniform Hypergraph}

Next, we evaluated Hyper-SAGNN using the non-$k$-uniform heterogeneous hypergraph. 
For the above four datasets, we decomposed each hyperedge into 3 pairwise edges and added them to the existing graph.
We trained our model to predict both the hyperedges and the edges  (i.e., non-hyperedges).
We then evaluated the performance for link prediction tasks for both the hyperedges and the edges. 
We also performed the node classification task following the same setting as above.
The results for link prediction are in Table~\ref{tab:link}.
Fig.~\ref{fig:perform_node} shows the results for the node classification task. 

\begin{table}[ht!]
\caption{Performance evaluation based on AUROC and AUPR for hyperedge/edge prediction. 
Methods with annotation (mix) represent Hyper-SAGNN trained with a mixture of edges and hyper-edges. 
Datasets marked with ``(2)'' represent the performance on pair-wise edge prediction (i.e., non-hyperedges).}
    \begin{center} \sf \footnotesize 
        \begin{tabular}{l|ll|ll|ll|ll}
        \toprule 
     &  \multicolumn{2}{c|}{\textbf{GPS}} &  \multicolumn{2}{c|}{\textbf{MovieLens}} &  \multicolumn{2}{c|}{\textbf{drug}} &  \multicolumn{2}{c}{\textbf{wordnet}} \\ 
    & AUC & AUPR & AUC & AUPR & AUC & AUPR & AUC & AUPR \\ \hline
     node2vec - mean & 0.563 & 0.191 & 0.562 & 0.197 & 0.670 & 0.246 & 0.608 & 0.213 \\ 
    node2vec - min & 0.570 & 0.185 & 0.539 & 0.186 & 0.684 & 0.258  & 0.575 & 0.200 \\ 
    DHNE  &0.910 &0.668 & 0.877 & 0.668 & 0.925 & 0.859 & 0.816 & 0.459 \\ 
    Hyper-SAGNN-E & \textbf{0.952} & \textbf{0.798} & 0.926 & 0.793 & \textbf{0.961} & \textbf{0.895} & \textbf{0.890} & 0.705\\ 
    Hyper-SAGNN-W  &0.922 & 0.722 & \textbf{0.930} & \textbf{0.810} & 0.955 & 0.892 & 0.880 & \textbf{0.706}\\
    Hyper-SAGNN-E (mix) &0.950 & 0.795 & 0.928 & 0.799 & 0.956 & 0.887 &0.881 & 0.694\\
   Hyper-SAGNN-W (mix) &0.920 & 0.720 & 0.929 & 0.811 & 0.950 & 0.889 &0.884 & 0.684 \\
   \bottomrule  
   \toprule
       &  \multicolumn{2}{c|}{\textbf{GPS (2)}} &  \multicolumn{2}{c|}{\textbf{MovieLens (2)}} &  \multicolumn{2}{c|}{\textbf{drug (2)}} &  \multicolumn{2}{c}{\textbf{wordnet (2)}} \\ 
    & AUC & AUPR & AUC & AUPR & AUC & AUPR & AUC & AUPR \\ \hline
   Hyper-SAGNN-E (mix) &0.921 &0.899 &0.971 & 0.967 &0.981 &0.973 &0.891 & 0.897\\
    Hyper-SAGNN-W (mix) &0.931 &0.910 &0.999 & 0.999 &0.999 &0.999 &0.923 & 0.916\\
\bottomrule 
        \end{tabular}
    \label{tab:link}
    \end{center}
\end{table}

We observed that Hyper-SAGNN can preserve the graph structure on different levels.
Compared to training the model with hyperedges only, including the edges into the training would not cause obvious changes in performance for hyperedge predictions (about a 1\% fluctuation for AUC/AUPR). 

\begin{figure}[ht!]
    \centering
    \includegraphics[width = 0.95\textwidth]{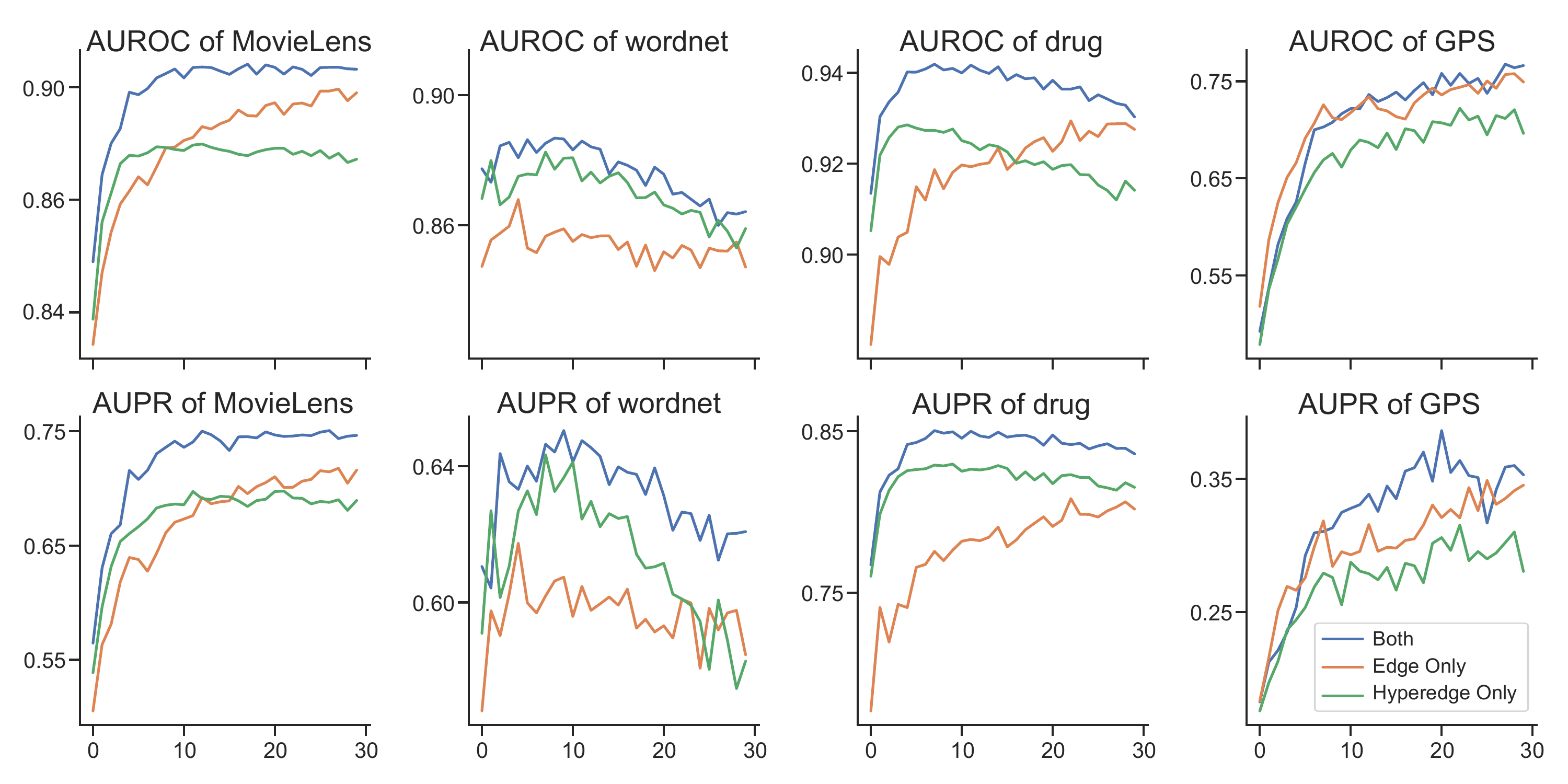}
    \caption{AUROC and AUPR scores of Hyper-SAGNN for hyperedge prediction on the downsampled dataset over training epochs.}
    \label{fig:perform_down}
\end{figure}

We then further assessed the model in a new evaluation setting where there are adequate edges but only a few hyperedges presented.
We asked whether the model can still achieve good performance on the hyperedge prediction based on this dataset.
This scenario is possible in the real-world applications especially when the dataset is combined from different sources. 
For example, in the drug dataset, it is possible that, in addition to the (user, drug, reaction) hyperedges, there are also extra edges that come from other sources, e.g., (drug, reaction) edges from the drug database, (user, drug) and (user, reaction) edges from the medical record. 
Here for each dataset that we tested, we used about 50\% of the edges and only 5\% of the hyperedges in the network to train the model. 
The results are in Fig.~\ref{fig:perform_down}.

When using only the edges to train the model, our method still achieves higher AUROC and AUPR score for hyperedge prediction as compared to node2vec (Table~\ref{tab:link}).
We found that when the model is trained with both the downsampled hyperedge dataset and the edge dataset, it would be able to reach higher performance or suffer less from overfitting than being trained with each of the datasets individually.
This demonstrates that our model can capture the consensus information on the graph structure across different sizes of hyperedges.

\subsection{Outsider Identification}
In addition to the standard link prediction and node classification, 
we further formulated a new task called ``outsider identification''. 
Previous methods such as DHNE can answer the question of whether a specific tuple of nodes $(v_1, v_2, ..., v_k)$ form a hyperedge. 
However, in many settings, we might also want to know the reason why this group of nodes will not form a hyperedge. 
We first define the outsider of a group of nodes as follows.
Node $v_i$ is the outsider of the node group $(v_1, v_2, ..., v_k)$ if it satisfies:
\begin{align}
    &\exists e \in E, (v_1,v_2,...,v_{i-1},...,v_{i+1},...,v_k) \in e \\
    & \nexists  e \in E, s.t.~\exists j \in \{1,2,..,k\}, j\neq i, (v_i, v_j) \in e
\end{align}

We speculated that Hyper-SAGNN can answer this question by analyzing the probability score $p_1$ to $p_k$ (defined in Eqn.~\ref{eq:p_i}). 
We assume that the node $v_i$ with the smallest $p_i$ would be the outsider.
We then set the evaluation as follows. 
We first trained the model as usual, but at the final stage, we replaced the average pooling layer with min pooling layer and fine-tuned the model for several epochs. 
We then fed the generated triplets with known outsider node into the trained model and calculated the top-$k$ accuracy of the outsider node matching the node with the smallest probability. 
Because this task is based on the prediction results of the hyperedges, we only tested on the dataset that achieves the best hyperedge prediction, i.e., the drug dataset.
We found that we have 81.9\% accuracy for the smallest probability and 95.3\% accuracy for the top-2 smallest probability. 
These results showed that by switching the pooling layer we would have better outsider identification accuracy (from 78.5\% to 81.9\%) with the cost of slightly decreased hyperedge prediction performance (AUC from 0.955 to 0.935).
This demonstrates that our model is able to accurately predict the outsider within the group even without further labeled information.
Moreover, the performance of outsider identification can be further improved if we include the cross-entropy between $p_i$ and the label of whether $v_i$ is an outsider for all applicable triplets in the loss term.
Together, these results demonstrate the advantage of Hyper-SAGNN in terms of the interpretability of hyperedge prediction. 

\subsection{Application to Single-cell Hi-C Datasets}

We next applied Hyper-SAGNN to the recently produced single-cell Hi-C (scHi-C) datasets~\citep{ramani2017massively,nagano2017cell}. 
Genome-wide mapping of chromatin interactions by Hi-C~\citep{lieberman2009comprehensive,rao20143d} has enabled comprehensive characterization of the 3D genome organization that reveals patterns of chromatin interactions between genomic loci. 
However, unlike bulk Hi-C data where signals are aggregated from cell populations, scHi-C provides unique information about chromatin interactions at single-cell resolution, thus allowing us to ascertain cell-to-cell variation of the 3D genome organization. 
Specifically, we propose that scHi-C makes it possible to model the cell-to-cell variation of chromatin interaction as a hyperedge, i.e., (cell, genomic locus, genomic locus). 
For the analysis of scHi-C, the most common strategy would be revealing the cell-to-cell variation by embedding the cells based on the contact matrix and then applying the clustering algorithms such as \textit{K}-means clustering or hierarchical clustering on the embedded vectors. 
We performed the following evaluation to assess the effectiveness of Hyper-SAGNN on learning the embeddings of cells by representing the scHi-C data as hypergraphs.

We tested Hyper-SAGNN on two datasets. 
The first one consists of scHi-C from four human cell lines: HAP1, GM12878, K562, and HeLa~\citep{ramani2017massively}. 
The second one includes the scHi-C that represents the cell cycle of the mouse embryonic stem cells~\citep{nagano2017cell}. 
We refer to the first dataset as ``Ramani et al. data'', and the second as ``Nagano et al. data'' for abbreviation. 

\begin{figure}[ht!]
    \centering
    \includegraphics[width = 0.98\textwidth]{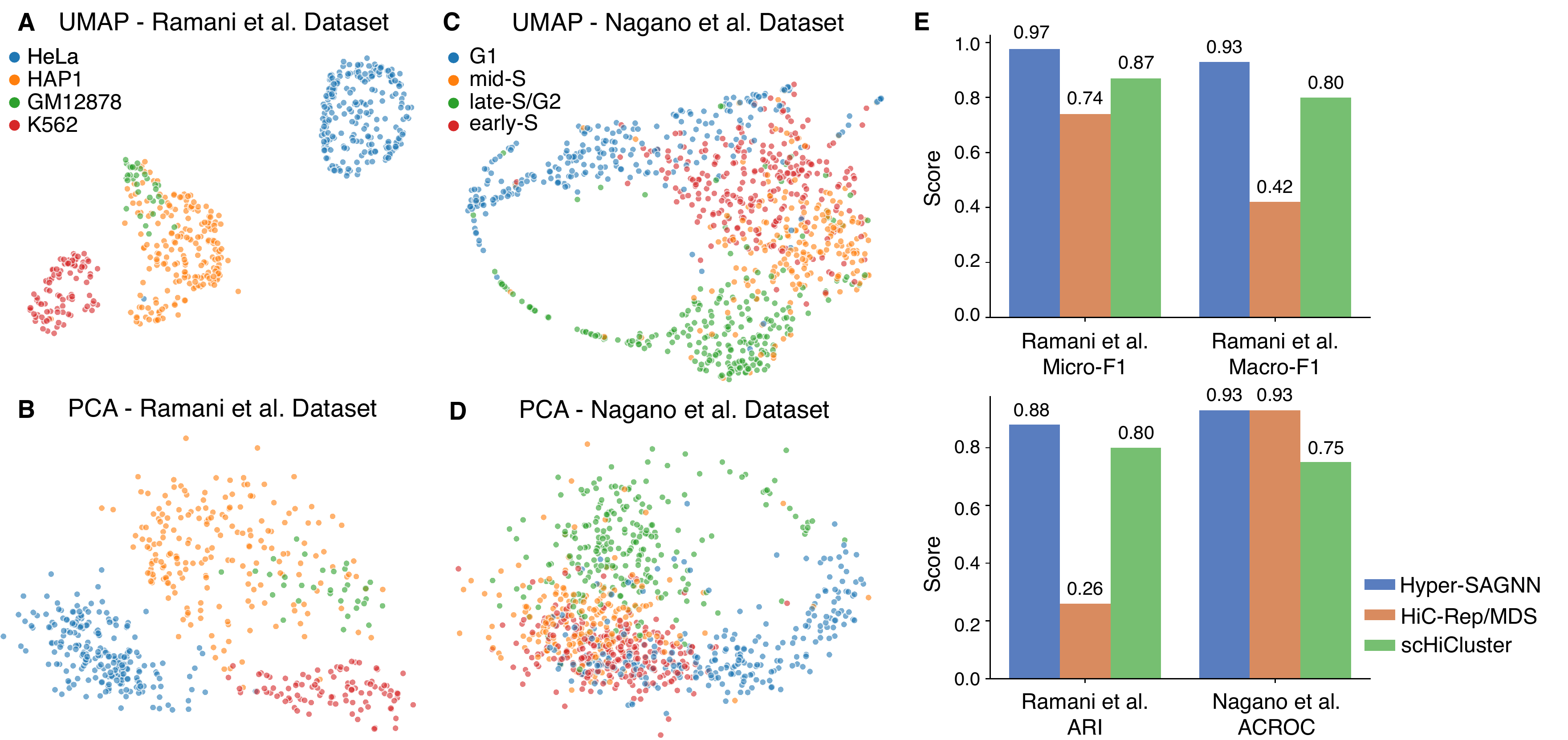}
    \caption{\textbf{(A) and (B):} Visualization of the learned embedding based on Hyper-SAGNN for the Ramani et al. data. 
    \textbf{(C) and (D):} Visualization of the learned embedding based on Hyper-SAGNN for the Nagano et al. data. 
    Embedding vectors are projected to two dimensional space using either UMAP or PCA.
    \textbf{(E):} Quantitative evaluation of the Hyper-SAGNN on two scHi-C datasets}
    \label{fig:vis_schics}
\end{figure}

We trained Hyper-SAGNN with the corresponding datasets. 
Due to the large average degrees of cell nodes, the random walk approach would take an extensive amount of time to sample the walks.
Thus, we only applied the encoder version of our method.
We visualize the learned embeddings by reducing them to 2 dimensions with PCA and UMAP~\citep{mcinnes2018umap} (Fig.~\ref{fig:vis_schics}A-D).

We quantified the effectiveness of the embeddings by applying \textit{K}-means clustering on the Ramani et al. data and evaluating with Adjusted Rand Index (ARI). 
In addition, we also assessed the effectiveness of the embeddings with a supervised scenario.
We used Logistic Regression as the classifier with 10\% of the cell as training samples and evaluated the multi-class classification task with Micro-F1 and Macro-F1.
We did not run \textit{K}-means clustering on the Nagano et al. data as it represents a state of continuous cell cycle which is not suitable for a clustering task. 
We instead used the metric ACROC (Average Circular ROC) developed in the HiCRep/MDS paper~\citep{liu2018unsupervised} to evaluate the performance of the three methods on the Nagano et al. data.
We compared the performance with two recently developed computational methods based on dimensionality reduction of the contact matrix, HiC-Rep/MDS~\citep{liu2018unsupervised} and scHiCluster~\citep{zhou2019robust}. 
Because Hyper-SAGNN is not a deterministic method for generating embeddings for scHi-C, we repeated the training process 5 times and averaged the score. 
All these results are in Fig.~\ref{fig:vis_schics}E.

For the Ramani et al. data (Fig.~\ref{fig:vis_schics}A-B), the visualization of the embedding vectors learned by Hyper-SAGNN exhibits clear patterns that cells with the same cell type are clustered together. 
Moreover, cell line HAP1, GM12878, and K562 are all blood-related cell lines, which are likely to be more similar to each other in terms of 3D genome organization as compared to HeLa.
Indeed, we observed that they are also closer to each other in the embedding space.
Quantitative results in Fig.~\ref{fig:vis_schics}E are consistent with the visualization as our method achieves the highest ARI, Micro-F1, Macro-F1 score among all three methods.
For the Nagano et al. data, as shown in Fig.~\ref{fig:vis_schics}C-D, we found that the embeddings exhibit a circular pattern that corresponds to the cell cycle.
Also, both HiC-Rep/MDS and Hyper-SAGNN achieve high ACROC score.
All these results demonstrated the effectiveness of representing the scHi-C datasets as hypergraphs using Hyper-SAGNN, which has great potential to provide insights into the cell-to-cell variation of higher-order genome organization.

\section{Conclusion}

In this work, we have developed a new graph neural network model called Hyper-SAGNN for the representation learning of general hypergraphs. 
The framework has the flexible ability to deal with homogeneous and heterogeneous, and uniform and non-uniform hypergraphs.
We demonstrated that Hyper-SAGNN is able to improve or match state-of-the-art performance for hypergraph representation learning while addressing the shortcomings of prior methods such as the inability to predict hyperedges for non-$k$-uniform heterogeneous hypergraphs. 
Hyper-SAGNN is computationally efficient as the size of input to the graph attention layer is bounded by the maximum hyperedge size as opposed to the number of first-order neighbors.

One potential improvement of Hyper-SAGNN as future work would be to allow information aggregation over all the first-order neighbors before calculating the static/dynamic embeddings for a node with additional computational cost.
With this design, the static embedding for a node would still satisfy our constraint that it is fixed for a known hypergraph with varying input tuples. 
This would allow us to incorporate previously developed methods on graphs, such as GraphSAGE~\citep{hamilton2017inductive} and GCN~\citep{kipf2016semi},
as well as methods designed for hypergraphs like HyperGCN~\citep{yadati2018hypergcn}
into this framework for better link prediction performance. 
Such improvement may also extend the application of Hyper-SAGNN to semi-supervised learning.

\section*{Acknowledgment}

J.M. acknowledges support from the National Institutes of Health Common Fund 4D Nucleome Program grant U54DK107965, National Institutes of Health grant R01HG007352, and National Science Foundation grant 1717205. 
Y.Z. (Yao Class, IIIS, Tsinghua University) contributed to this work as a visiting undergraduate student at Carnegie Mellon University during summer 2019.


\bibliographystyle{abbrvnat}
\bibliography{refs}

\begin{thebibliography}{26}
\providecommand{\natexlab}[1]{#1}
\providecommand{\url}[1]{\texttt{#1}}
\expandafter\ifx\csname urlstyle\endcsname\relax
  \providecommand{\doi}[1]{doi: #1}\else
  \providecommand{\doi}{doi: \begingroup \urlstyle{rm}\Url}\fi

\bibitem[Bordes et~al.(2013)Bordes, Usunier, Garcia-Duran, Weston, and
  Yakhnenko]{datasetwordnet}
A.~Bordes, N.~Usunier, A.~Garcia-Duran, J.~Weston, and O.~Yakhnenko.
\newblock Translating embeddings for modeling multi-relational data.
\newblock In \emph{Advances in Neural Information Processing Systems}, pages
  2787--2795, 2013.

\bibitem[Feng et~al.(2018)Feng, He, Liu, Nie, and Chua]{feng2018learning}
F.~Feng, X.~He, Y.~Liu, L.~Nie, and T.-S. Chua.
\newblock Learning on partial-order hypergraphs.
\newblock In \emph{Proceedings of the 2018 World Wide Web Conference}, pages
  1523--1532, 2018.

\bibitem[Grover and Leskovec(2016)]{grover2016node2vec}
A.~Grover and J.~Leskovec.
\newblock node2vec: Scalable feature learning for networks.
\newblock In \emph{Proceedings of the 22nd ACM SIGKDD International Conference
  on Knowledge Discovery and Data Mining}, pages 855--864. ACM, 2016.

\bibitem[Gui et~al.(2016)Gui, Liu, Tao, Jiang, Norick, and Han]{gui2016large}
H.~Gui, J.~Liu, F.~Tao, M.~Jiang, B.~Norick, and J.~Han.
\newblock Large-scale embedding learning in heterogeneous event data.
\newblock In \emph{2016 IEEE 16th International Conference on Data Mining
  (ICDM)}, pages 907--912. IEEE, 2016.

\bibitem[Hamilton et~al.(2017{\natexlab{a}})Hamilton, Ying, and
  Leskovec]{hamilton2017inductive}
W.~Hamilton, Z.~Ying, and J.~Leskovec.
\newblock Inductive representation learning on large graphs.
\newblock In \emph{Advances in Neural Information Processing Systems}, pages
  1024--1034, 2017{\natexlab{a}}.

\bibitem[Hamilton et~al.(2017{\natexlab{b}})Hamilton, Ying, and
  Leskovec]{hamilton2017representation}
W.~L. Hamilton, R.~Ying, and J.~Leskovec.
\newblock Representation learning on graphs: Methods and applications.
\newblock \emph{arXiv preprint arXiv:1709.05584}, 2017{\natexlab{b}}.

\bibitem[Harper and Konstan(2015)]{datasetMovieLens}
F.~M. Harper and J.~A. Konstan.
\newblock The movielens datasets: History and context.
\newblock \emph{ACM Trans. Interact. Intell. Syst.}, 5\penalty0 (4):\penalty0
  19:1--19:19, Dec. 2015.
\newblock ISSN 2160-6455.

\bibitem[Huang et~al.(2019)Huang, Chen, Ye, Wu, Zheng, and
  Ling]{huang2019hyper2vec}
J.~Huang, C.~Chen, F.~Ye, J.~Wu, Z.~Zheng, and G.~Ling.
\newblock Hyper2vec: Biased random walk for hyper-network embedding.
\newblock In \emph{International Conference on Database Systems for Advanced
  Applications}, pages 273--277. Springer, 2019.

\bibitem[Kipf and Welling(2016)]{kipf2016semi}
T.~N. Kipf and M.~Welling.
\newblock Semi-supervised classification with graph convolutional networks.
\newblock \emph{arXiv preprint arXiv:1609.02907}, 2016.

\bibitem[Lieberman-Aiden et~al.(2009)Lieberman-Aiden, Van~Berkum, Williams,
  Imakaev, Ragoczy, Telling, Amit, Lajoie, Sabo, Dorschner,
  et~al.]{lieberman2009comprehensive}
E.~Lieberman-Aiden, N.~L. Van~Berkum, L.~Williams, M.~Imakaev, T.~Ragoczy,
  A.~Telling, I.~Amit, B.~R. Lajoie, P.~J. Sabo, M.~O. Dorschner, et~al.
\newblock Comprehensive mapping of long-range interactions reveals folding
  principles of the human genome.
\newblock \emph{Science}, 326\penalty0 (5950):\penalty0 289--293, 2009.

\bibitem[Liu et~al.(2018)Liu, Lin, Yard{\i}mc{\i}, and
  Noble]{liu2018unsupervised}
J.~Liu, D.~Lin, G.~G. Yard{\i}mc{\i}, and W.~S. Noble.
\newblock Unsupervised embedding of single-cell hi-c data.
\newblock \emph{Bioinformatics}, 34\penalty0 (13):\penalty0 i96--i104, 2018.

\bibitem[McInnes et~al.(2018)McInnes, Healy, and Melville]{mcinnes2018umap}
L.~McInnes, J.~Healy, and J.~Melville.
\newblock Umap: Uniform manifold approximation and projection for dimension
  reduction.
\newblock \emph{arXiv preprint arXiv:1802.03426}, 2018.

\bibitem[Mikolov et~al.(2013{\natexlab{a}})Mikolov, Chen, Corrado, and
  Dean]{mikolov2013word2vec}
T.~Mikolov, K.~Chen, G.~Corrado, and J.~Dean.
\newblock Efficient estimation of word representations in vector space.
\newblock \emph{arXiv preprint arXiv:1301.3781}, 2013{\natexlab{a}}.

\bibitem[Mikolov et~al.(2013{\natexlab{b}})Mikolov, Sutskever, Chen, Corrado,
  and Dean]{mikolov2013skipgram}
T.~Mikolov, I.~Sutskever, K.~Chen, G.~S. Corrado, and J.~Dean.
\newblock Distributed representations of words and phrases and their
  compositionality.
\newblock In \emph{Advances in Neural Information Processing Systems}, pages
  3111--3119, 2013{\natexlab{b}}.

\bibitem[Nagano et~al.(2017)Nagano, Lubling, V{\'a}rnai, Dudley, Leung, Baran,
  Cohen, Wingett, Fraser, and Tanay]{nagano2017cell}
T.~Nagano, Y.~Lubling, C.~V{\'a}rnai, C.~Dudley, W.~Leung, Y.~Baran, N.~M.
  Cohen, S.~Wingett, P.~Fraser, and A.~Tanay.
\newblock Cell-cycle dynamics of chromosomal organization at single-cell
  resolution.
\newblock \emph{Nature}, 547\penalty0 (7661):\penalty0 61, 2017.

\bibitem[Perozzi et~al.(2014)Perozzi, Al-Rfou, and Skiena]{perozzi2014deepwalk}
B.~Perozzi, R.~Al-Rfou, and S.~Skiena.
\newblock Deepwalk: Online learning of social representations.
\newblock In \emph{Proceedings of the 20th ACM SIGKDD International Conference
  on Knowledge Discovery and Data Mining}, pages 701--710. ACM, 2014.

\bibitem[Ramani et~al.(2017)Ramani, Deng, Qiu, Gunderson, Steemers, Disteche,
  Noble, Duan, and Shendure]{ramani2017massively}
V.~Ramani, X.~Deng, R.~Qiu, K.~L. Gunderson, F.~J. Steemers, C.~M. Disteche,
  W.~S. Noble, Z.~Duan, and J.~Shendure.
\newblock Massively multiplex single-cell hi-c.
\newblock \emph{Nature Methods}, 14\penalty0 (3):\penalty0 263, 2017.

\bibitem[Rao et~al.(2014)Rao, Huntley, Durand, Stamenova, Bochkov, Robinson,
  Sanborn, Machol, Omer, Lander, et~al.]{rao20143d}
S.~S. Rao, M.~H. Huntley, N.~C. Durand, E.~K. Stamenova, I.~D. Bochkov, J.~T.
  Robinson, A.~L. Sanborn, I.~Machol, A.~D. Omer, E.~S. Lander, et~al.
\newblock A 3d map of the human genome at kilobase resolution reveals
  principles of chromatin looping.
\newblock \emph{Cell}, 159\penalty0 (7):\penalty0 1665--1680, 2014.

\bibitem[Sun et~al.(2008)Sun, Ji, and Ye]{sun2008hypergraph}
L.~Sun, S.~Ji, and J.~Ye.
\newblock Hypergraph spectral learning for multi-label classification.
\newblock In \emph{Proceedings of the 14th ACM SIGKDD international Conference
  on Knowledge Discovery and Data Mining}, pages 668--676. ACM, 2008.

\bibitem[Tu et~al.(2018)Tu, Cui, Wang, Wang, and Zhu]{tu2018structural}
K.~Tu, P.~Cui, X.~Wang, F.~Wang, and W.~Zhu.
\newblock Structural deep embedding for hyper-networks.
\newblock In \emph{Thirty-Second AAAI Conference on Artificial Intelligence},
  2018.

\bibitem[Vaswani et~al.(2017)Vaswani, Shazeer, Parmar, Uszkoreit, Jones, Gomez,
  Kaiser, and Polosukhin]{vaswani2017attention}
A.~Vaswani, N.~Shazeer, N.~Parmar, J.~Uszkoreit, L.~Jones, A.~N. Gomez,
  {\L}.~Kaiser, and I.~Polosukhin.
\newblock Attention is all you need.
\newblock In \emph{Advances in Neural Information Processing Systems}, pages
  5998--6008, 2017.

\bibitem[Veli{\v{c}}kovi{\'c} et~al.(2017)Veli{\v{c}}kovi{\'c}, Cucurull,
  Casanova, Romero, Lio, and Bengio]{velivckovic2017graph}
P.~Veli{\v{c}}kovi{\'c}, G.~Cucurull, A.~Casanova, A.~Romero, P.~Lio, and
  Y.~Bengio.
\newblock Graph attention networks.
\newblock \emph{arXiv preprint arXiv:1710.10903}, 2017.

\bibitem[Yadati et~al.(2018)Yadati, Nimishakavi, Yadav, Louis, and
  Talukdar]{yadati2018hypergcn}
N.~Yadati, M.~Nimishakavi, P.~Yadav, A.~Louis, and P.~Talukdar.
\newblock Hypergcn: Hypergraph convolutional networks for semi-supervised
  classification.
\newblock \emph{arXiv preprint arXiv:1809.02589}, 2018.

\bibitem[Zheng et~al.(2010)Zheng, Cao, Zheng, Xie, and Yang]{datasetGPS}
V.~W. Zheng, B.~Cao, Y.~Zheng, X.~Xie, and Q.~Yang.
\newblock Collaborative filtering meets mobile recommendation: A user-centered
  approach.
\newblock In \emph{Proceedings of the Twenty-Fourth AAAI Conference on
  Artificial Intelligence}, AAAI'10, pages 236--241. AAAI Press, 2010.

\bibitem[Zhou et~al.(2007)Zhou, Huang, and Sch{\"o}lkopf]{zhou2007learning}
D.~Zhou, J.~Huang, and B.~Sch{\"o}lkopf.
\newblock Learning with hypergraphs: Clustering, classification, and embedding.
\newblock In \emph{Advances in Neural Information Processing Systems}, pages
  1601--1608, 2007.

\bibitem[Zhou et~al.(2019)Zhou, Ma, Chen, Cheng, Bao, Peng, Sejnowski, Dixon,
  and Ecker]{zhou2019robust}
J.~Zhou, J.~Ma, Y.~Chen, C.~Cheng, B.~Bao, J.~Peng, T.~J. Sejnowski, J.~R.
  Dixon, and J.~R. Ecker.
\newblock Robust single-cell hi-c clustering by convolution-and
  random-walk--based imputation.
\newblock \emph{Proceedings of the National Academy of Sciences}, page
  201901423, 2019.

\end{thebibliography}

\clearpage
\appendix

\setcounter{page}{1}
\renewcommand\thefigure{A\arabic{figure}}
\setcounter{figure}{0}    
\renewcommand{\thetable}{A\arabic{table}}
\setcounter{table}{0} 

\section{Appendix}

\subsection{Comparison of Hyper-SAGNN with Its Variants}
\label{app:variants}

As mentioned above, unlike the standard GAT model, we exclude the $\alpha_{ii}$ term in the self-attention mechanism. 
To test whether this constraint would improve or reduce the model's ability to learn, we implemented a variant of our model (referred to as variant type I) by including this term. 
Also, as mentioned in the Method section, another potential variant of our model would be directly using the $\vec{d}^{*}_i$ to calculate the probability score $p^{*}_i$. 
We refer to this variant as variant type II. 
For variant type II, on node classification task, since it does not have a static embedding, we used $W_v^T x_i$.
The rest of the parameters and structure of the neural network remain the same. 

\begin{figure}
    \centering
    \includegraphics[width = 0.6\textwidth]{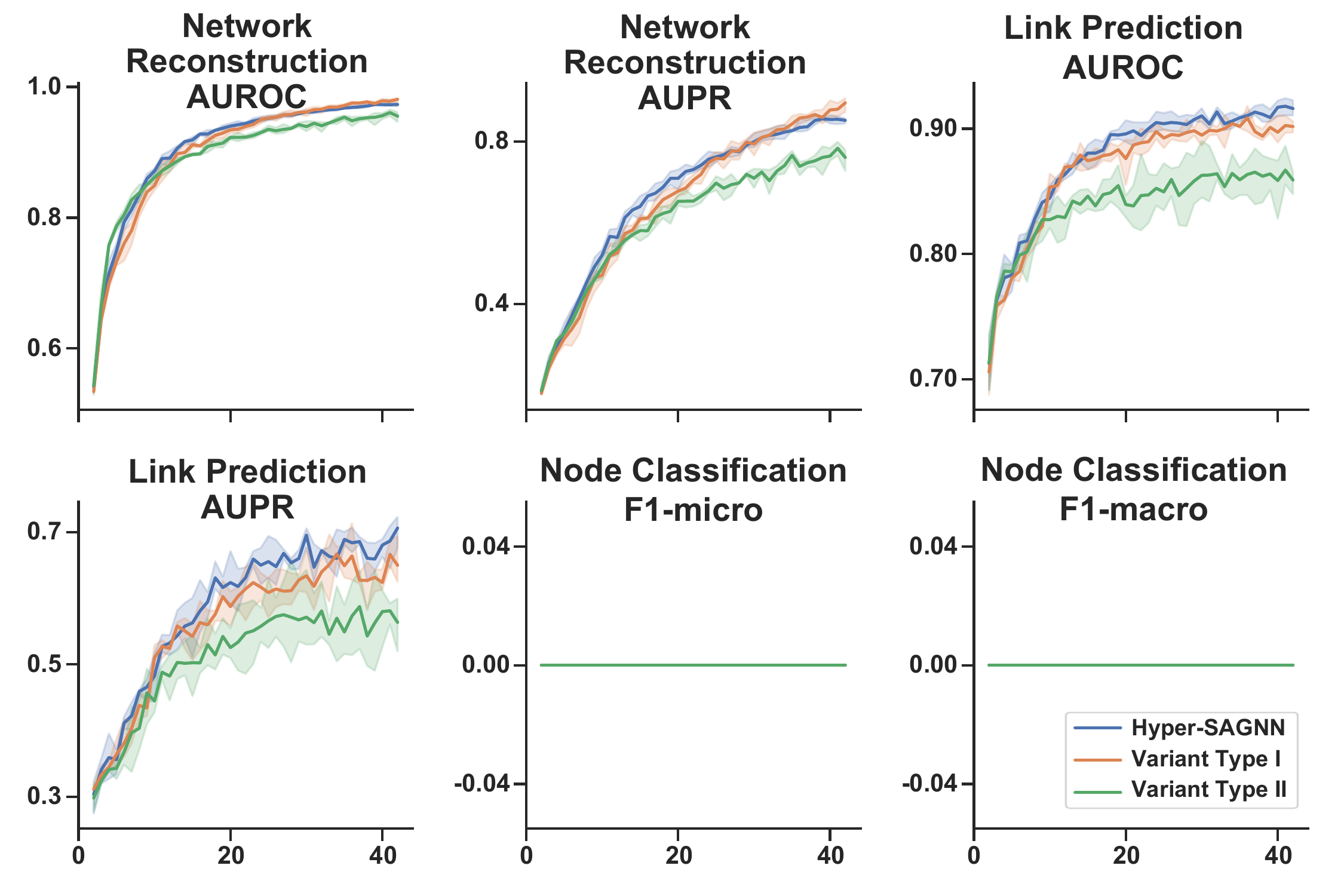}
    \caption{Performance comparison of Hyper-SAGNN -- Walk and Variant Type I, II (GPS)}
    \label{fig:perform_GPS1}
\end{figure}

\begin{figure}
    \centering
    \includegraphics[width = 0.6\textwidth]{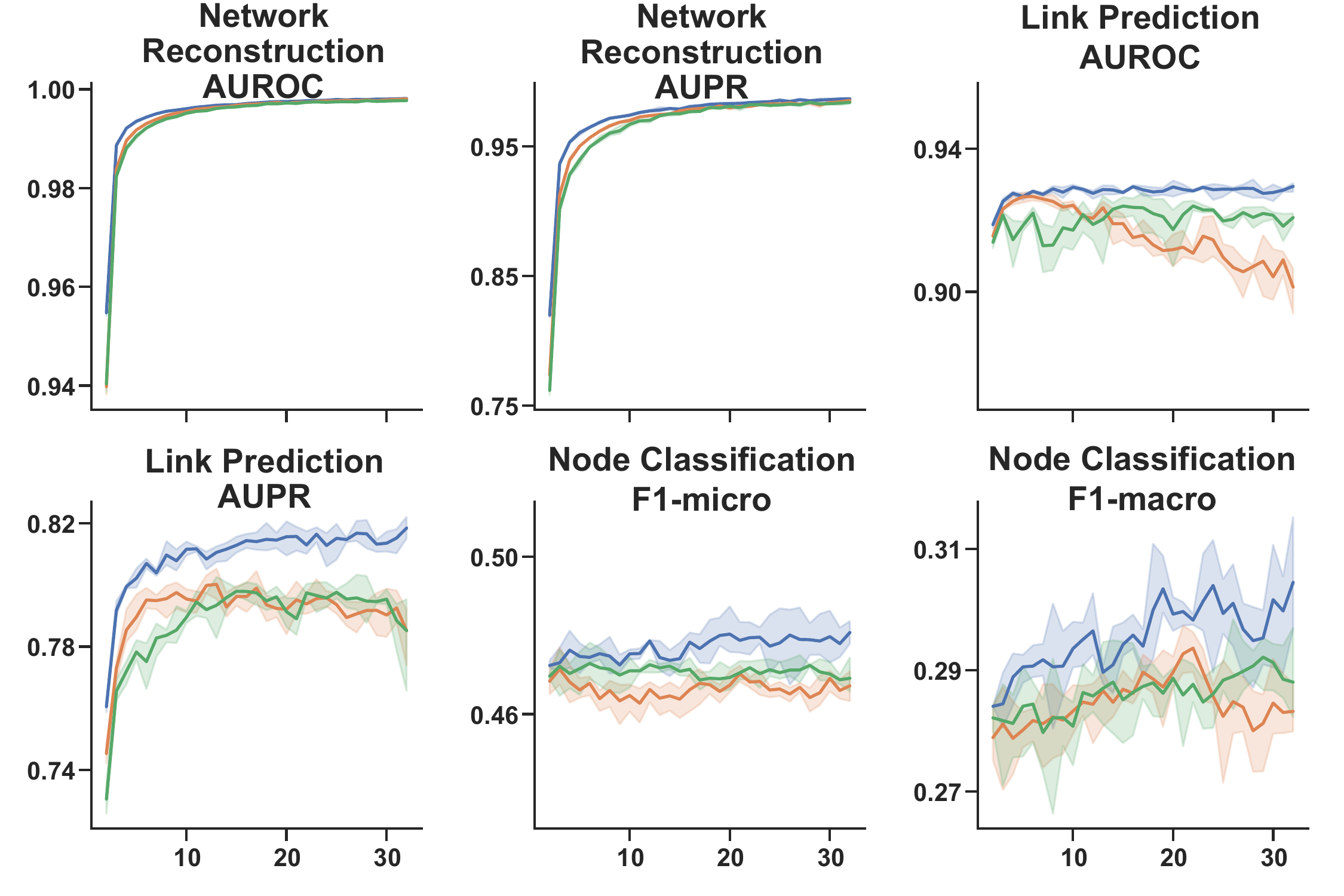}
    \caption{Performance comparison of Hyper-SAGNN -- Walk and Variant Type I, II (MovieLens)}
    \label{fig:perform_movie1}
\end{figure}

\begin{figure}
    \centering
    \includegraphics[width = 0.6\textwidth]{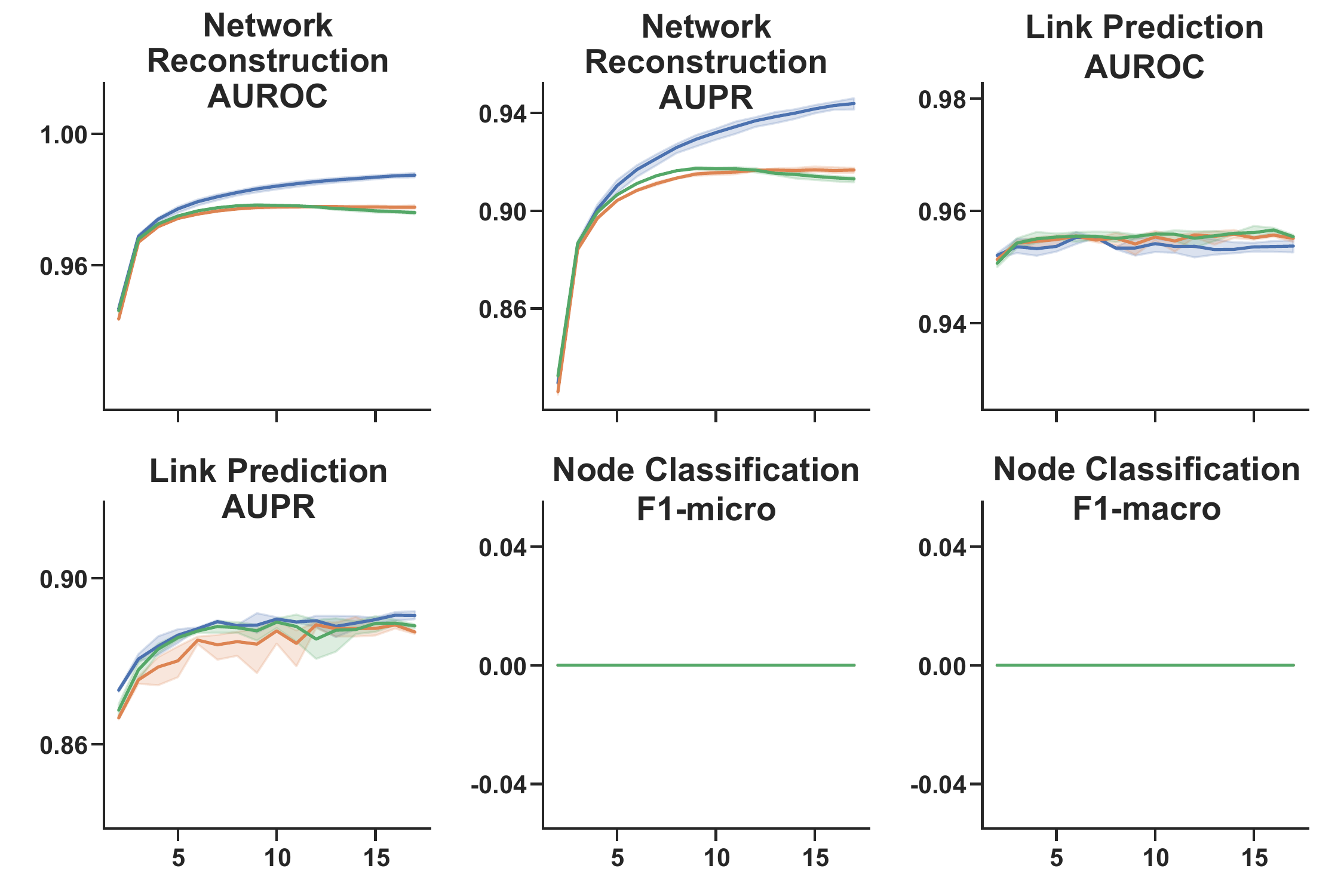}
    \caption{Performance comparison of Hyper-SAGNN -- Walk and Variant Type I, II (drug)}
    \label{fig:perform_drug1}
\end{figure}

\begin{figure}
    \centering
    \includegraphics[width = 0.6\textwidth]{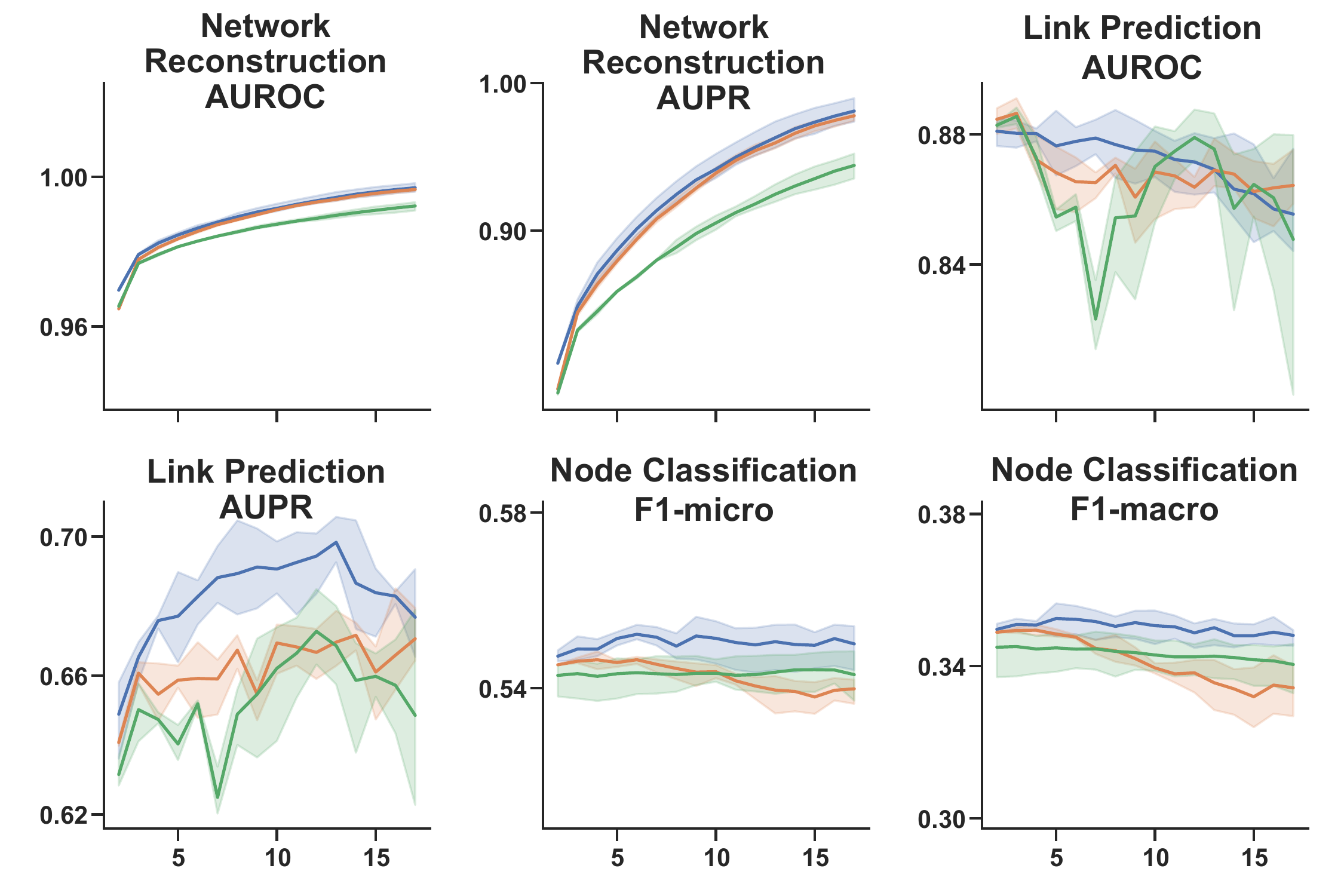}
    \caption{Performance comparison of Hyper-SAGNN -- Walk and Variant Type I, II (wordnet)}
    \label{fig:perform_wordnet1}
\end{figure}

\begin{figure}
    \centering
    \includegraphics[width = 0.6\textwidth]{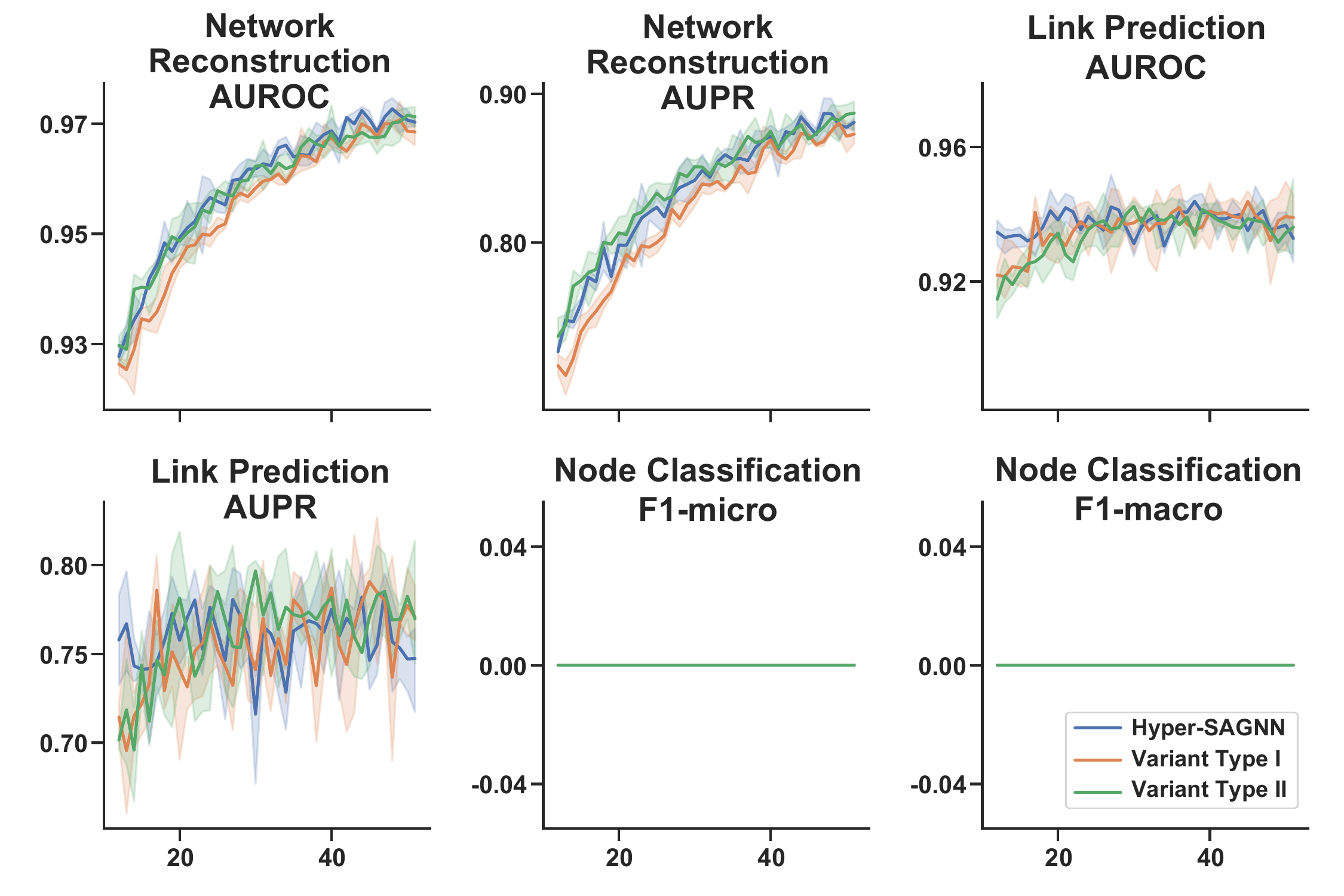}
    \caption{Performance comparison of Hyper-SAGNN -- Encoder and Variant Type I, II (GPS)}
    \label{fig:perform_GPS2}
\end{figure}

\begin{figure}
    \centering
    \includegraphics[width = 0.6\textwidth]{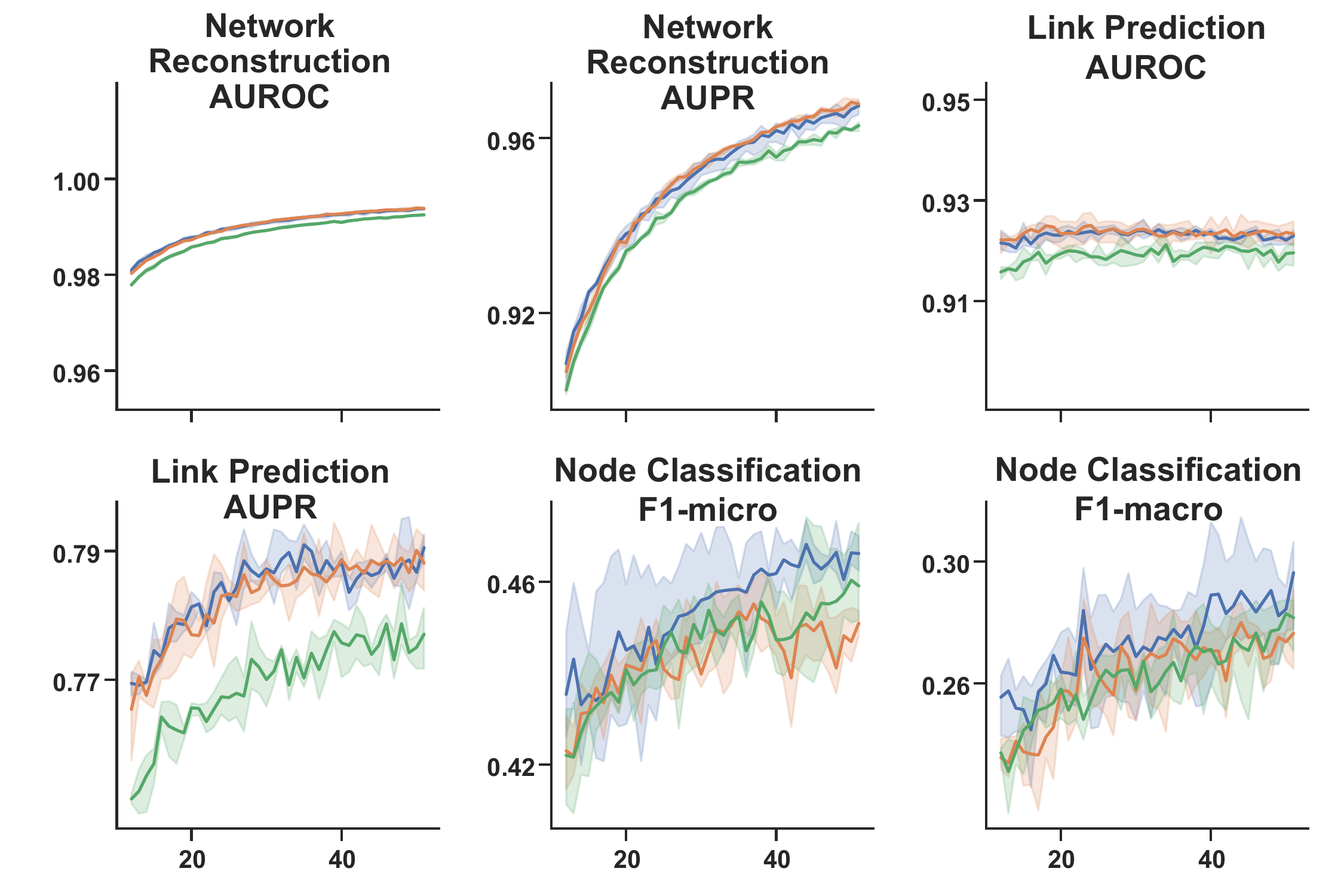}
    \caption{Performance comparison of Hyper-SAGNN -- Encoder and Variant Type I, II (MovieLens)}
    \label{fig:perform_movie2}
\end{figure}

\begin{figure}
    \centering
    \includegraphics[width = 0.6\textwidth]{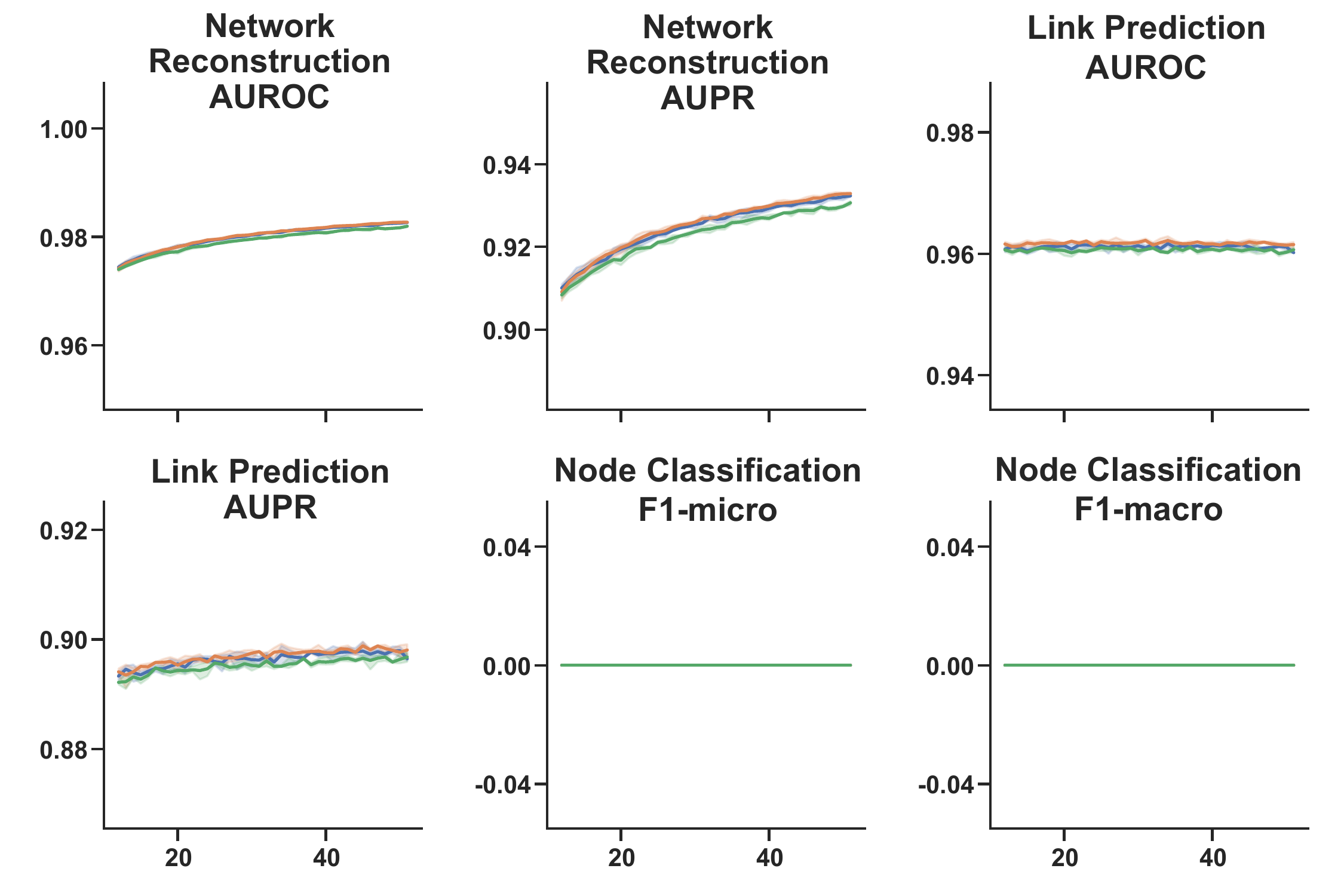}
    \caption{Performance comparison of Hyper-SAGNN -- Encoder and Variant Type I, II (drug)}
    \label{fig:perform_drug2}
\end{figure}

\begin{figure}
    \centering
    \includegraphics[width = 0.6\textwidth]{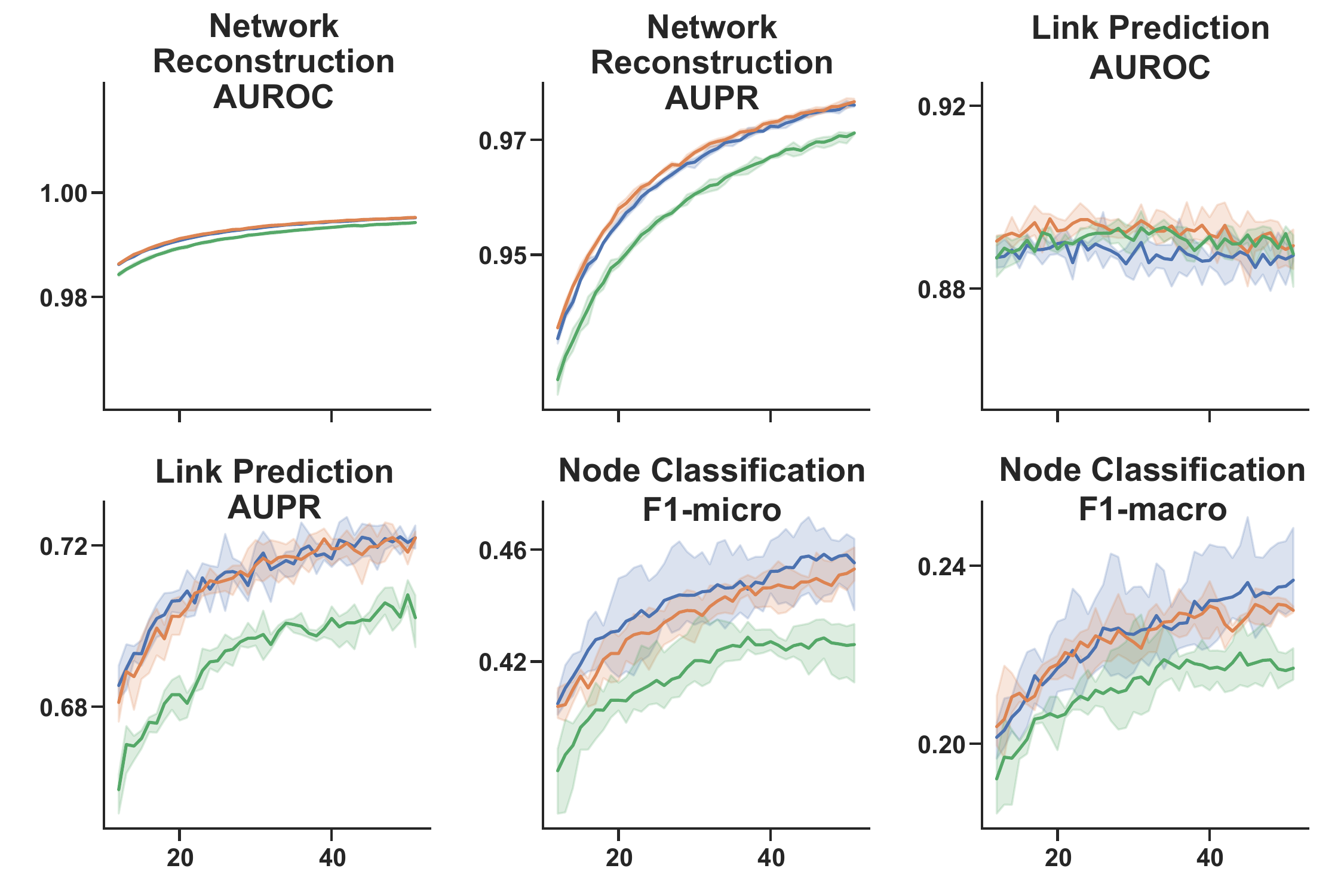}
    \caption{Performance comparison of Hyper-SAGNN -- Encoder and Variant Type I, II (wordnet)}
    \label{fig:perform_wordnet2}
\end{figure}

We then compared the performance of Hyper-SAGNN and two variants in terms of AUC and AUPR values for network reconstruction task and hyperedge link prediction task on the following four datasets: MovieLens, wordnet, drug, and GPS. 
We also compared the performance in terms of the Micro F1 score and Macro F1 score on the node classification task on the MovieLens and the wordnet dataset.
For the MovieLens dataset, we used 90\% nodes as training data while for wordnet, we used 1\% of the nodes as training data.
All the evaluation setup is the same as described in the main text. 
To avoid the effect of randomness from the neural network training, we repeated the training process for each experiment five times and made the line plot of the score versus the epoch number. 
To illustrate the differences more clearly, we started the plot at epoch 3 for the random walk based approach and epoch 12 for the encoder based approach.
The performance of the model using the random walk based approach is shown in Fig.~\ref{fig:perform_GPS1} to Fig.~\ref{fig:perform_wordnet1}. 
The performance of the model using the encoder based approach is shown in Fig.~\ref{fig:perform_GPS2} to Fig.~\ref{fig:perform_wordnet2}.

For models with the random walk based approach, Hyper-SAGNN is the best in terms of all metrics for the GPS, MovieLens, and wordnet dataset.
On the drug dataset, Hyper-SAGNN achieves higher AUROC and AUPR score on the network reconstruction task than two variants, but slightly lower AUROC score for the link prediction task (less than 0.5\%). 

For models with the encoder based approach, the advantage is not that obvious. 
All 3 methods achieve similar performance in terms of all metrics for the GPS and the drug dataset. 
For the MovieLens and wordnet dataset, Hyper-SAGNN performs similar to variant type I, higher than variant type II on the network reconstruction and link prediction task. 
However, our model achieves slightly higher accuracy on the node classification task than variant type I.

Therefore, these evaluations demonstrated that the choice of the structure of Hyper-SAGNN can achieve higher or at least comparable performance than the two potential variants over multiple tasks on multiple datasets.

\clearpage

\end{document}